\pdfoutput=1

\documentclass[11pt]{article}

\usepackage[preprint]{acl}
\usepackage[table]{xcolor}

\usepackage{times}
\usepackage{latexsym}

\usepackage[T1]{fontenc}

\usepackage[utf8]{inputenc}

\usepackage{microtype}

\usepackage{inconsolata}

\usepackage{graphicx}
\usepackage[strings]{underscore}
\usepackage{amsmath}
\usepackage{enumitem}
\usepackage{arydshln}
\usepackage{xcolor}
\usepackage{amsfonts}
\usepackage{algorithm}      
\usepackage{algpseudocode}  
\usepackage{footnote}
\usepackage{hyperref}
\usepackage{array}       
\usepackage{multirow}    

\title{MADS: Model-Aware Diverse Core Set Selection for Instruction Tuning}

\author{
 \textbf{Yi Bai\textsuperscript{1}},
 \textbf{Wenhao Zhang\textsuperscript{1}},
 \textbf{Yao Chen\textsuperscript{2}},
 \textbf{Jiao Xue\textsuperscript{2}},
 \textbf{Zhumin Chen\textsuperscript{1},
 \textbf{Pengjie Ren\textsuperscript{1}}\thanks{Corresponding author.}}
\\
 \textsuperscript{1}Shandong University, Qingdao, China
 \textsuperscript{2}Inspurcloud, Jinan, China
\\
 \texttt{\{202235147, zhangwenhao\}@mail.sdu.edu.cn,} \\
 \texttt{\{chenyao, xuejiao02\}@inspur.com,} \\
 \texttt{\{chenzhumin,renpengjie\}@sdu.edu.cn}
}

\begin{document}
\maketitle

\begin{abstract}
Instruction fine-tuning is employed to enhance the instruction-following ability of large language models (LLMs).
As the amount of instruction fine-tuning data increases, selecting the optimal core set becomes particularly important.
However, ensuring the diversity of the core set remains a significant challenge.
Existing methods predominantly distinguish different training data based on the text features themselves, decoupled from LLMs' own understanding and representation of the data.
To address this issue, we propose a Model-Aware Diverse Core Set Selection method, which distinguishes data features based on the neural activation states during LLM inference.
This approach serves as an efficient instantiation of coverage-based selection using model-intrinsic activation features to ensure the diversity in the core set.
We extensively evaluate our method on six benchmarks that cover five distinct tasks.
In our method, the core set selected by the 3B-parameter LLM performs effectively when utilized to fine-tune larger models with 7B, 8B, and 13B parameters.
Experimental results on the Alpaca-GPT4 dataset, which comprises 52K instruction–response pairs, show that the core set, sized at 15\% of the original dataset and selected by Llama-3.2-3B-Instruct, achieves an average improvement of 2.5\% when fine-tuning four larger base models compared with training on the full dataset.
The experimental results demonstrate that our method enhances model performance on multiple downstream tasks while reducing data requirements.
\end{abstract}

\section{Introduction}
\label{sec:intro}
With the rapid advancement of artificial intelligence technology, large language models (LLMs) such as GPT-4~\citep{achiam2023gpt}, Mistral~\citep{jiang2023mistral}, Llama 3~\citep{dubey2024llama}, and Qwen~\citep{bai2023qwen} have demonstrated outstanding performance in various tasks through large-scale training data and powerful computing capabilities.
During the pre-training phase, LLMs are trained on large corpora to acquire general language knowledge and logical reasoning skills.
The fine-tuning phase aims to enhance the model's ability to follow instructions and align with human preferences~\citep{sanhmultitask, ouyang2022training}.
Therefore, carefully curated fine-tuning data is crucial for optimizing model performance.

Early approaches to obtaining high-quality instruction data involved collecting instruction-response pairs through crowdsourcing or leveraging powerful LLMs to generate instruction datasets~\citep{sanhmultitask, taori2023stanford, wang2023self}.
However, as the volume of available training data increases, using all of it for fine-tuning becomes impractical.
Moreover, some studies indicate that increasing the amount of instruction data does not always enhance model performance~\citep{shi2024continual, wu2024continual}.
\citet{zhou2024lima} found that a small but well-chosen instruction dataset can outperform a larger one.
Some studies have attempted to select data based on instance-level quality~\citep{cao2023instruction, pang2024improving, li2024quantity, zhang2025best}. 
In contrast, our approach focuses on dataset-level features, such as diversity and coverage, which have been shown to play a more significant role in data selection than individual data quality~\citep{xia2024rethinking}.
Methods based on diversity and coverage select new data points that are most distinct from the already selected ones. 
Subsets selected by these methods often achieve performance superior or comparable to that achieved using the full dataset during fine-tuning~\citep{chen2023maybe,lu2023instag,das2024deft,shao-etal-2024-balanced}.

However, measuring and ensuring the diversity and coverage of data remains a significant challenge.
We categorize existing methods for selecting instruction data into two types: data-aware methods and model-aware methods.
(1) Data-aware methods use pre-trained language models like BERT~\citep{devlin-etal-2019-bert} to extract data representations, ensuring uniform data distribution through k-means clustering or directly assigning category labels using powerful LLMs~\citep{chen2023maybe, lu2023instag, das2024deft, shao-etal-2024-balanced}.
When selecting diverse data, these methods do not fully utilize the internal representations of LLMs to guide the data selection.
(2) Model-aware methods use the LLM to be fine-tuned or already fine-tuned to assess the necessity of each data instance, selecting data beneficial to the model~\citep{li2023one, liu2024selectit, li2024quantity, zhang2025best, hu2025donod, aligncot2023, selfrefineit2024}.
These methods tend to select data that is challenging or learnable for the current model, rather than diversity.

Motivated by the aforementioned challenges, we propose a novel method, \underline{M}odel-\underline{A}ware \underline{D}iverse Core Set \underline{S}election for Instruction Tuning \footnote{\href{https://anonymous.4open.science/r/MADS-5711/}{https://anonymous.4open.science/r/MADS-5711/}}, which leverages the internal representations of LLMs to select a core set that is both diverse and highly representative.
The core idea of MADS is to use the neuron activation states generated by LLMs during inference as representations of instruction data to select a diverse data subset.
Previous research has shown that LLM neurons exhibit different activation states for different input data features~\citep{elhage2022superposition, bricken2023monosemanticity, bills2023language, cunningham2023saeinterpretable, luo2025inversescope, helff2025activationreasoning, shafran2025decomposing}.

A potential concern is whether activation-based representations can reliably capture semantic features, given that individual neurons in LLMs are known to be polysemantic---a single neuron may respond to multiple unrelated concepts~\citep{elhage2022superposition}. However, recent interpretability research reveals that neural networks encode independent features through \emph{linear combinations of neurons} rather than individual neurons~\citep{bricken2023monosemanticity}. This insight motivates our approach: instead of tracking which individual neurons are activated, we record the \emph{set of neurons} that are strongly co-activated by each instruction as its activation tag. This group-level representation naturally captures the compositional feature structure of LLMs and mitigates the polysemanticity issue.

To empirically validate the correlation between activation tags and semantic features, we conduct a preliminary experiment using 1,000 randomly selected instructions from each of five domains. Through PCA visualization and pairwise similarity analysis of activation tags across multiple layers of Llama-3.2-3B-Instruct, we find that same-domain instructions share significantly more activation tags than cross-domain pairs, confirming that activation tags capture domain-specific semantic features. The full empirical analysis, including visualization figures and detailed domain discussion, is presented in Section~\ref{sec:preliminary_validation}.
Inspired by the correlation between neuron activation patterns and data features, we are the first to use neuron activation states of LLMs as data representations for diverse instruction data selection.
Specifically, MADS calculates all neuron activations in the original dataset and filters a subset covering all activation patterns as the core set.
During core set selection, we prioritize complex instructions that activate a richer set of neurons, as complex instructions more effectively enhance the comprehension and reasoning capabilities of LLMs~\citep{lu2023instag}.
Compared with existing methods, MADS extracts model-level data features, ensuring the selected data subset has superior diversity, coverage, and complexity.
Moreover, neuron activation can be extracted in a single inference, eliminating the need for additional training and reducing computational and time costs.

We conduct extensive experiments on instruction-following benchmarks, demonstrating that fine-tuning LLMs with data selected by MADS achieves superior instruction-following performance compared to existing methods.
We also perform further analysis to verify the coverage and robustness of our method.
Our contributions can be summarized as follows:
\begin{itemize}[topsep=0pt, partopsep=0pt, itemindent=0pt]
    \setlength{\itemsep}{0pt}
    \setlength{\parsep}{0pt}
    \setlength{\parskip}{1pt}
    \item We introduce a novel model-aware diverse instruction data selection method, utilizing neuron activation states of LLMs for the first time to achieve diverse and complex instruction data selection.
    \item Our method extracts data representations in a single inference step, without the need for manual data category definition or gradient calculation, improving the efficiency of instruction data selection.
    \item Our experiments with Alpaca show that our method improves LLMs across various tasks using just 15\% of the data, outperforming others with more significant and balanced improvements.
\end{itemize}

\section{Related Work}
\subsection{Instruction Data Selection}

{
\setlength{\parindent}{0cm}
\textbf{Data-aware methods.}
}
Data-aware methods focus on the quality, diversity, and importance of the instruction during data selection~\citep{qin2024unleashing}.
To ensure the quality of the instruction, \citet{cao2023instruction} designed a metric system to evaluate the text quality such as lexical diversity and dialogue coherence. 
Moreover, \citet{xu2023rethinking, liu2023makes, pang2024improving} leverage powerful LLMs like GPT-4 to measure data quality based on various aspects such as instruction complexity and response accuracy.
In terms of diversity, the most common approach involves utilizing pre-trained language models like BERT to embed data into a high-dimensional space, followed by clustering with methods such as k-means or k-center to select a subset with a uniform distribution~\citep{chen2023maybe, das2024deft, shao-etal-2024-balanced}.
\citet{lu2023instag} utilizes GPT-4 to classify instructions, thereby selecting a data subset that covers multiple categories.
Importance is also considered a criterion, referring to the difficulty level of an instruction-response pair for LLMs.
To identify hard instructions, \citet{du2023mods} employs a reward model to assess whether LLMs could generate correct responses to the given instructions.
\citet{song2024iterselecttune} trains BERT as a classifier to distinguish between easy and hard instructions.
These methods typically rely on additional models to classify data, making the data selection process independent of the internal representations of LLMs.

{
\setlength{\parindent}{0cm}
\textbf{Model-aware methods.}
}
Model-aware methods typically treat the data to be selected as input and use probability distributions, loss, gradients, or other model-related metrics generated by LLMs for data selection~\citep{zhang2025best, dai2025improving, zhao2025beyond, zhou2025davir}.
For instance, \citet{li2024quantity} compares the loss generated by LLMs with and without instruction context to estimate the difficulty of the instruction.
\citet{hu2025donod} developed two model-parameter-based metrics to filter out noisy, unlearnable, and generalization-impairing samples.
Similarly, \citet{liu2024selectit} uses different grain uncertainties of LLMs to improve the accuracy of the data. 
In addition to these methods that only require LLMs to perform inference, there are approaches that use gradients from backpropagation as data selection criteria~\citep{joaquin2024in2core, pan2024g}.
\citet{yang2024smalltolarge} leverages training trajectories to select mathematical data.
These model-aware methods tend to select data that is more difficult for LLMs rather than more diverse data.

\subsection{Core Set Selection}
The goal of core set selection is to select a subset from all training data such that the model trained on this subset achieves performance similar to that of the model trained on the full dataset. 
Core set selection has deep roots in classical machine learning. 
Early theoretical work established foundational algorithms based on geometric methods such as k-median and k-means clustering~\citep{har2005smaller}, while subsequent research extended these ideas to logistic regression~\citep{munteanu2018coresets}, gradient-based selection~\citep{mirzasoleiman2020coresets}, and deep learning scenarios~\citep{paul2021deep}. 
Closely related to core set selection is Active Learning, which iteratively selects the most informative samples for labeling~\citep{settles2009active, sener2018active}. 
Methods such as uncertainty sampling and diversity-based selection from Active Learning share similar goals with core set selection in reducing data requirements while maintaining model performance~\citep{ash2019deep, margatina2021active}.
However, existing methods either are computationally intensive~\citep{xia2024less}, making them infeasible for large-scale LLM data selection, or rely on predefined concepts to categorize data for diversity. In contrast, our method leverages the LLM's own activation states to efficiently distinguish data types and ensure diversity.

\section{Methodology}
\subsection{Problem Formulation}
Given an initial full training dataset, $D_{full}$, which contains $n$ instruction-response pairs, our task is to select a subset $D_{c}$ from $D_{full}$ such that $|D_{c}| \ll |D_{full}|$.

The objective of our method is to leverage the neuron activation states of LLMs to select a core subset $D_{c}$ that can maximally cover the features present in $D_{full}$.
This core subset $D_{c}$ is designed to significantly reduce the amount of training data required for fine-tuning while enhancing downstream performance of LLMs.

\subsection{Overview}
\begin{figure*}[t]
    \centering
    \includegraphics[width=0.95\linewidth]{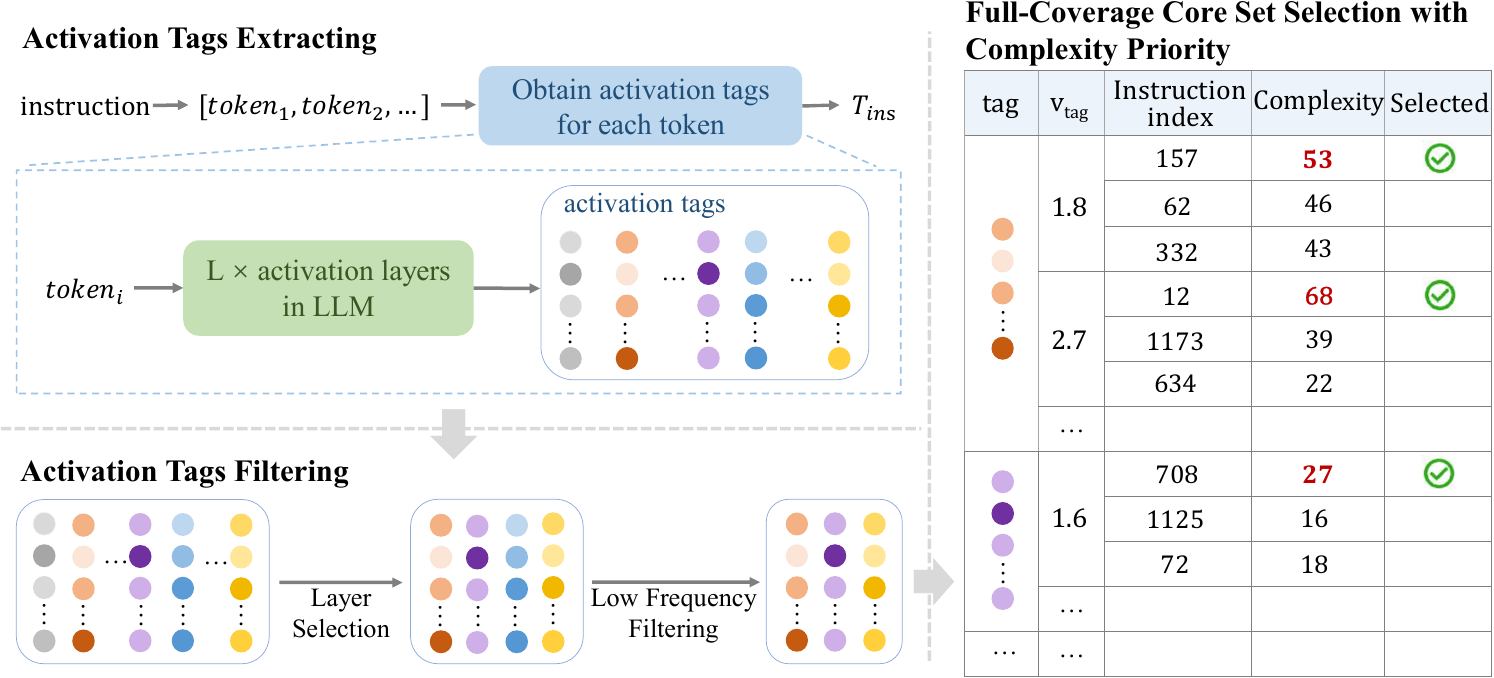}
    \caption {The overall framework of MADS, consisting of three stages.
    (1) \textbf{Activation Tags Extracting}: Instructions are fed into the LLM, and neuron activation values exceeding threshold 1 are extracted from each MLP layer and converted into activation tags.
    (2) \textbf{Activation Tags Filtering}: Representative layers are selected based on the proportion of strongly activated neurons, and low-frequency activation tags are filtered out to reduce noise.
    (3) \textbf{Full-Coverage Core Set Selection with Complexity Priority}: For each unique activation pattern $(tag, v_{tag})$, the instruction with the highest complexity (most distinct activation tags) is selected to ensure comprehensive coverage while prioritizing complex instructions.
    }
    \label{fig:MADS}
\end{figure*}

Previous studies on the interpretability of LLMs have demonstrated that the activation states generated during inference can represent the features of input data~\citep{elhage2022superposition, bricken2023monosemanticity, bills2023language}.
In particular, \citet{bricken2023monosemanticity} propose that a neural network represents features of the data by assigning each feature its own linear combination of neurons, such that a corresponding set of neurons activates when the feature is present.
Inspired by this, we define the neurons activated by the training data during inference as their tags, which represent the features in the data.
Based on these activation tags, we perform core set selection by covering as many tags as possible, thereby ensuring a diverse set of features in the core set.
Our method comprises three steps: extracting activation tags, filtering activation tags, and sampling maximum complexity activation tags.
Figure~\ref{fig:MADS} provides an overview of our method.

\subsection{Activation Tags Extracting}
First, we derive the activation tags of the instructions in $D_{full}$.
We utilize a fine-tuned LLM to extract the output of the activation function for all instructions in each Multilayer Perceptron (MLP) layer during inference and convert these outputs into activation tags.

Before inference, instructions are encoded into multiple tokens $\{token_1, \ldots, token_N\}$.
These tokens are sequentially fed into the LLM with $L$ layers, each containing an activation function in the MLP submodule.
For $token_i$, we extract the output of the activation function from each of the $L$ layers:
\begin{equation}
    \label{eq:llm_activation}
    \mathbf{a}_{i}^{1}, \mathbf{a}_{i}^{2}, \ldots, \mathbf{a}_{i}^{L} = \rm{LLM_{act}}(token_i)
\end{equation}
where $\mathbf{a}_{i}^{l}$ is the output of the activation function for $token_i$ at the $l$-th layer, with $\mathbf{a}_{i}^{l} \in \mathbb{R}^{d}$, and $d$ denotes the dimension of the output of the LLM activation function.

We then transform $\{\mathbf{a}_{i}^{1}, \mathbf{a}_{i}^{2}, \ldots, \mathbf{a}_{i}^{L}\}$ into activation tags that reflect the features of $token_i$.
Each feature corresponds to a linear combination of neurons that are strongly activated.
The activation level of neurons indicates the model's "confidence" that some feature is present, while weak activations may sometimes be erroneous~\citep{elhage2022superposition, bricken2023monosemanticity}.
Therefore, we filter out neurons with high activation levels as activation tags. 
Referring to experimental results from previous studies~\citep{bricken2023monosemanticity} concerning the relationship between activation levels and their corresponding text, we set the filtering threshold to 1. 
We provide a series of examples in the Appendix~\ref{appendix:appendix} to illustrate this point.

To further validate this threshold, we analyze 39 billion activation values in our experiments and find that 99.70\% of them are less than or equal to 1, indicating that only approximately 0.30\% of activations exceed this threshold. 
Referring to recent work~\citep{shafran2025decomposing} that retains the top 1\% of activations as strong activations in Llama 3.1 models, we believe that setting this threshold to 1 is sufficient to filter out erroneous activations.

Let $\text{Idx}(\mathbf{a}_{i}^{l})$ denote the set of dimensions in $\mathbf{a}_{i}^{l}$ with activation value greater than 1:
\begin{equation}
    \label{eq:Idx}
    \text{Idx}(\mathbf{a}_{i}^{l}) = \{ j \mid \mathbf{a}_{i}^{l}[j] > 1 \}
\end{equation}
where $j=1,2,...,d$ is the index of the dimension in $\mathbf{a}_{i}^{l}$.

Considering the varying degrees of activation in the dimensions of $\text{Idx}(a_{i}^{l})$, we retain this information by sorting the dimensions in descending order according to their corresponding activation values and using them as the activation tag $tag_{i}^{l}$.
\begin{equation}
    \label{eq:sorted_Idx}
    tag_{i}^{l} = \text{Sort}(\text{Idx}(\mathbf{a}_{i}^{l}), \text{key} = \mathbf{a}_{i}^{l}[j])
\end{equation}

In addition to extracting activation tags, we also retain their corresponding activation values, as inspired by~\citet{bricken2023monosemanticity}, we hypothesize that activation values are correlated with data characteristics.
The activation tag value $v_{tag_{i}^{l}}$ is defined as the maximum activation value among the neurons indexed by $tag_{i}^{l}$:
\begin{equation}
    \label{eq:tag_val}
  v_{tag_{i}^{l}} = \max_{j \in tag_{i}^{l}} \mathbf{a}_{i}^{l}[j]
\end{equation}
To investigate our hypothesis, we analyze 260K activation tags (with $v_{tag_{i}^{l}}$ rounded to one decimal place, using standard rounding, i.e., round half up) and find that 98.14\% have only one unique activation value, while 1.86\% have multiple values.
A sampling study on the 1.86\% of tags with multiple activation values reveals a large number of samples exhibiting the following phenomenon: while instructions sharing the same activation tag possess common characteristics, those with different activation values within the tag exhibit finer-grained distinctions, with partial examples presented in Table~\ref{tab:activation_value_cases}.
This observation motivates us to incorporate activation values into the core set selection process to achieve finer-grained data partitioning.

Based on the above, the activation tags of $token_i$ are defined as $T(token_i) = \{tag_{i}^{1},tag_{i}^{2},...,tag_{i}^{L}\}$, where the activation tags of the instruction $ins$ are the union of all tokens it contains:
\begin{equation}
    \label{eq:tag_input}
    \begin{aligned}
        T(ins)&=T(token_1) \cup T(token_2) \cup \cdots \cup T(token_N) \\
        &=\left\{ \text{tag}_{i}^{l} \mid i = 1, 2, \ldots, N; \, l = 1, 2, \ldots, L \right\}
    \end{aligned}
\end{equation}

\subsection{Activation Tags Filtering}
In the previous section, we introduce the activation tags of the instruction $T(ins)$.
Here, considering the following two factors: 
(1) the activation tags between adjacent layers may have significant overlap, and
(2) different layers of LLMs capture different features~\citep{belinkov2018evaluating, peters2018dissecting, blevins2018deep}, we propose filtering the activation tags.
The filtering process consists of two steps: selecting representative layers based on activation patterns, and removing low-frequency activation tags to reduce noise.

\begin{figure}[t]
    \centering
    \includegraphics[width=0.95\columnwidth]{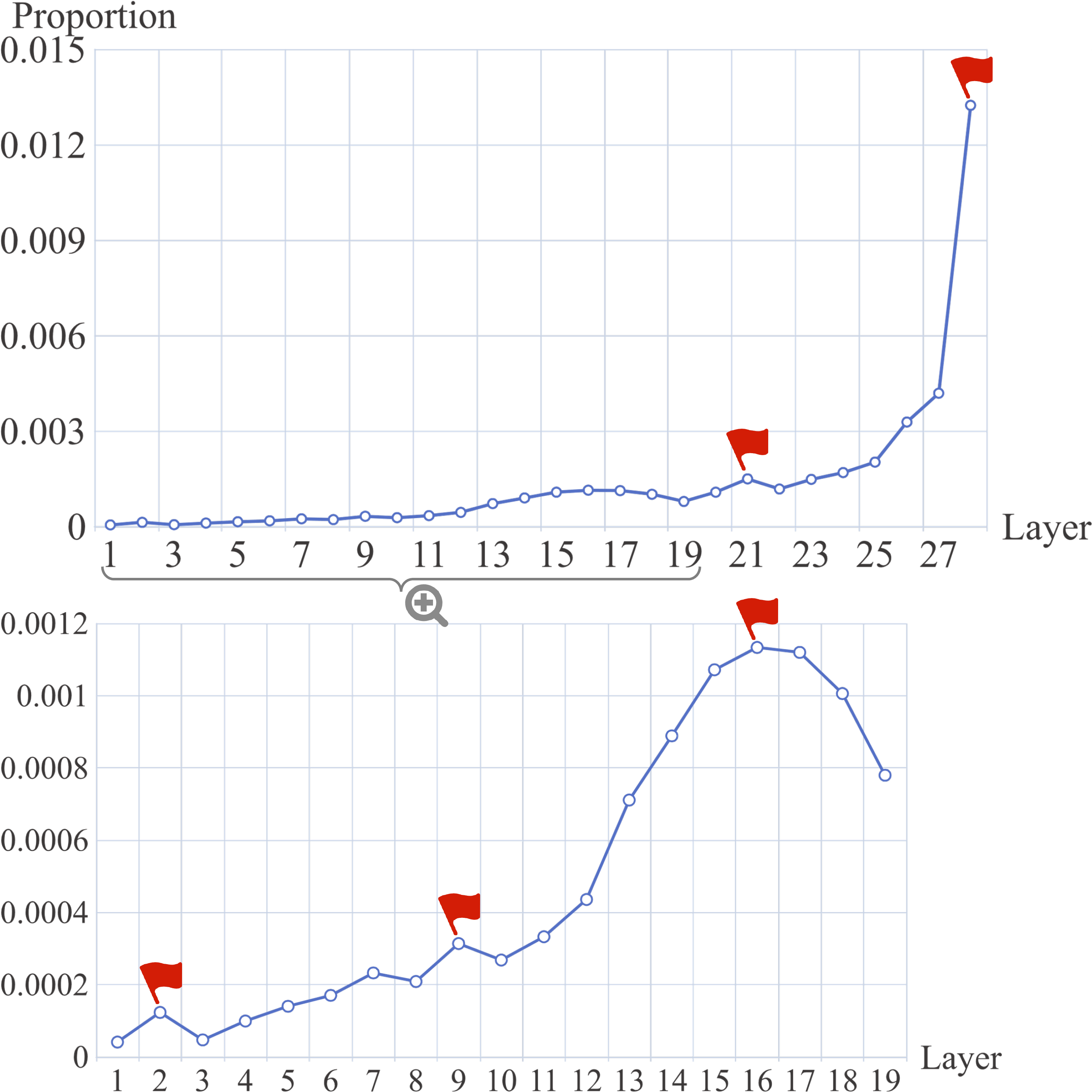}
    \caption{The proportion of strongly activated neurons (activation value $> 1$) in each layer of Llama-3.2-3B-Instruct.
    The x-axis represents the layer index (1--28), and the y-axis represents the average proportion of neurons with activation values exceeding 1.
    Red flags indicate selected layers (2nd, 9th, 16th, 21st, and 28th), corresponding to local maxima in the wave-like pattern.
    The 7th layer is excluded despite being a local maximum because it is adjacent to the 9th layer, which has a higher proportion value.
    This selection strategy ensures even distribution across network depth while prioritizing information-rich layers.}
    \label{fig:layer_select}
\end{figure}

\textbf{Step 1: Layer Selection Based on Activation Patterns.}
Assuming we need to select $M$ layers from $L$ layers as final activation tags, our principle is to select as evenly as possible.
Following this principle, we further design a specific layer selection strategy, which selects $M$ layers with a high proportion of strongly activated neurons.
For the $l$-th layer, we first extract the activation vectors of all instructions in $D_{full}$.
We then calculate the proportion of activation values greater than 1 within each activation vector.
Averaging these proportion values gives us the overall proportion of strongly activated neurons in the $l$-th layer.
Formally, for an instruction $ins$ at the $l$-th layer, let $\mathbf{a}_{ins}^{l} \in \mathbb{R}^{d}$ denote its activation vector. The proportion of strongly activated neurons for this instruction at layer $l$ is defined as:
\begin{equation}
    \label{eq:proportion}
    p_l(ins) = \frac{|\{j \mid \mathbf{a}_{ins}^{l}[j] > 1, j = 1, \ldots, d\}|}{d}
\end{equation}
The overall proportion for layer $l$ is computed as the average across all instructions:
\begin{equation}
    \label{eq:avg_proportion}
    p_l = \frac{1}{|D_{full}|} \sum_{ins \in D_{full}} p_l(ins)
\end{equation}
This process is repeated for each layer in the LLM, allowing us to track the proportion of strongly activated neurons across layers.
As depicted in Figure~\ref{fig:layer_select}, this proportion increases in a wave-like pattern starting from the first layer.
A higher proportion may indicate that the layer contains more information, with the peak of the wave representing the local maximum proportion of strongly activated neurons.
We select layers corresponding to these peaks as candidates for final activation tags. However, when multiple peaks occur in close proximity (i.e., separated by only one layer), selecting all of them would violate our principle of even distribution across the network depth. In such cases, we retain only the peak with the higher proportion value. 
Consequently, we select the layer corresponding to these peaks for final activation tags, denoted as $SL=\{SL_1, SL_2, ..., SL_M\}$.

\textbf{Step 2: Low-Frequency Activation Tags Filtering.}
After selecting $M$ layers, we further filter out the activation tags that appear infrequently.
This is because activation tags with very low frequency lack generality and can even be noise.
For each unique activation tag $tag$, we define its frequency as:
\begin{equation}
    \label{eq:frequency}
    f(tag) = |\{ins \in D_{full} \mid tag \in T(ins)\}|
\end{equation}
which counts the number of instructions in $D_{full}$ that contain the activation tag $tag$.
To filter out low-frequency activation tags, we first set a basic frequency filtering threshold $\theta_{base}$.
However, because the number of distinct activation tags varies significantly between lower and higher layers, it is unreasonable to set the same threshold for all layers.
To address this issue, we propose calculating the filtering weight for each layer based on the number of distinct activation tags contained in each layer.
For all activation tags of the instructions in $D_{full}$, we calculate the number of distinct activation tags contained in the $M$ layers we selected, denoted as $TN = \{TN_1, TN_2, ..., TN_M\}$.
Then, the filtering weight for the $l$-th layer is:
\begin{equation}
    \label{eq:layer_weight}
    w_{l} = \frac{max(TN_1, TN_2, ..., TN_M)}{TN_{l}} 
\end{equation}

We calculate the filtering threshold of the $l$-th layer:
\begin{equation}
    \label{eq:layer_threshold}
    \theta_l = \theta_{base} \times w_{l}
\end{equation}

Then, we filter out the low-frequency activation tags in the $l$-th layer with frequencies lower than the corresponding threshold $\theta_l$, for all selected layers $l \in SL$, to obtain the final activation tags of the instructions.
The activation tags of the instructions after filtering are:
\begin{equation}
    \label{eq:tag_filter}
    T^f(ins) = \{\text{tag}_{i}^{l} \mid l \in SL, \, \text{tag}_{i}^{l} \notin Tags_{f}\},
\end{equation}

where $Tags_{f}$ $= \{tag \mid f(tag) < \theta_l\}$ is the set of low-frequency activation tags to be filtered out.

\begin{algorithm*}[h]
  \caption{Full-Coverage Core Set Selection with Complexity Priority}
  \label{alg:core-set}
  \textbf{Input:} Full dataset $D_{full}$, filtered activation tags $T^f(ins)$ for each instruction, activation values $v_{tag}$ for each tag  \\
  \textbf{Output:} Core subset $D_c$
  
  \begin{algorithmic}[1]
  \State \textbf{// Step 1: Building Activation Pattern Mapping}
  \State Initialize $\mathcal{P} \gets \emptyset$ \Comment{Set of all activation patterns}
  \State Initialize $\mathcal{I} \gets \emptyset$ \Comment{Mapping from patterns to instructions}
  \For{each $ins \in D_{full}$}
      \For{each $\text{tag} \in T^f(ins)$}
          \State $v_{tag} \gets \text{activation value of } tag \text{ in } ins$
          \State $v_{tag} \gets \text{Round}(v_{tag}, 1)$ \Comment{Standard rounding (round half up) to one decimal place, enabling grouping of tags with similar activation intensity}
          \State $\mathcal{P} \gets \mathcal{P} \cup \{(tag, v_{tag})\}$ \Comment{Add activation pattern}
          \State Append $ins$ to $\mathcal{I}(tag, v_{tag})$ \Comment{Build pattern-to-instructions mapping}
      \EndFor
  \EndFor
  
  \State \textbf{// Step 2: Complexity-Based Representative Selection}
  \State Compute complexity $C(ins) = |T^f(ins)|$ for each $ins \in D_{full}$
  \State Initialize $D_c \gets \emptyset$
  
  \For{each $(tag, v_{tag}) \in \mathcal{P}$} \Comment{Iterate over each unique activation pattern}
      \State $\text{candidates} \gets \mathcal{I}(tag, v_{tag})$
      \State $ins^* \gets \arg\max_{ins \in \text{candidates}} C(ins)$ \Comment{Select most complex instruction}
      \If{$ins^* \notin D_c$}
          \State $D_c \gets D_c \cup \{ins^*\}$ \Comment{Ensure pattern coverage}
      \EndIf
  \EndFor
  
  \State \Return $D_c$
\end{algorithmic}
\end{algorithm*}

\subsection{Full-Coverage Core Set Selection with Complexity Priority}
In this stage, we describe how to select the core set $D_c$ from $D_{full}$ based on the filtered activation tags. Our goal is to ensure that $D_c$ covers all activation patterns in $D_{full}$ while prioritizing complex instructions. The selection process consists of two steps: (1) building a mapping from activation patterns to instructions, and (2) selecting representative instructions based on complexity.

\textbf{Step 1: Building Activation Pattern Mapping.}
For each instruction, we extract not only the activation tags but also their corresponding activation values. We use $(tag, v_{tag})$ pairs to represent distinct activation patterns because the same neuron may exhibit different activation intensities for different features.

For each instruction $ins$, we define its activation-pattern set as:
\begin{equation}
  \label{eq:pattern_per_ins}
  P^f(ins) = \{(tag, v_{tag}) \mid tag \in T^f(ins)\}
\end{equation}

The set of all distinct activation patterns is defined as the union of the activation-patterns of each instruction in $D_{full}$:
\begin{equation}
    \label{eq:pattern_set}
  \mathcal{P} = \bigcup_{ins \in D_{full}} P^f(ins)
\end{equation}
For each activation pattern $(tag, v_{tag}) \in \mathcal{P}$, we build a mapping to the set of instructions containing this pattern:
\begin{equation}
    \label{eq:pattern_mapping}
  \mathcal{I}(tag, v_{tag}) = \{ins \in D_{full} \mid (tag, v_{tag}) \in P^f(ins)\}
\end{equation}

\textbf{Step 2: Complexity-Based Representative Selection.}
For each instruction $ins$, we define its complexity $C(ins)$ as the number of distinct activation tags it contains:
\begin{equation}
    \label{eq:complexity}
    C(ins) = |T^f(ins)|
\end{equation}
Instructions with higher complexity values are considered more complex and are preferred as representatives, as they can potentially cover more activation patterns.

For each activation pattern $(tag, v_{tag}) \in \mathcal{P}$, we select the instruction with the highest complexity from $\mathcal{I}(tag, v_{tag})$ as the representative:
\begin{equation}
    \label{eq:representative}
    ins^* = \arg\max_{ins \in \mathcal{I}(tag, v_{tag})} C(ins)
\end{equation}
The final core set $D_c$ is the union of all selected representatives:
\begin{equation}
    \label{eq:core_set}
    D_c = \bigcup_{(tag, v_{tag}) \in \mathcal{P}} \{ins^*\}
\end{equation}

The detailed algorithm is presented in Algorithm~\ref{alg:core-set}.

\section{Experiments}
\subsection{Datasets}

{
\setlength{\parindent}{0cm}
\textbf{Training Datasets}
}
We utilize the Alpaca-GPT4 dataset~\citep{peng2023instruction}, which comprises 52,002 instruction-response pairs. The responses are generated by GPT-4 model~\citep{achiam2023gpt}, resulting in higher data quality compared to the original Alpaca dataset~\citep{taori2023stanford} from Stanford University.
WizardLM dataset~\citep{xu2024wizardlm} leverages the Evol-Instruct algorithm to generate high-quality instruction data.
Specifically, we used its WizardLM-7b subset comprising 70,000 instruction-response pairs.

{
\setlength{\parindent}{0cm}
\textbf{Evaluation Datasets}
}
To ensure a comprehensive and unbiased evaluation, we used 6 evaluation datasets covering 5 distinct tasks.

\begin{itemize}[itemsep=0pt,topsep=0pt]
    \item Factual knowledge: We use the Massive Multitask Language Understanding dataset (MMLU~\citep{hendrycks2020measuring}) to assess this ability. 
    MMLU assesses the ability of LLMs to understand factual knowledge, covering knowledge from 57 disciplines including STEM, humanities, and social sciences.
    \item Math Reasoning: We assess this ability using the Grade School Math dataset (GSM~\citep{cobbe2021training}). 
    GSM evaluates the mathematical ability of the LLM, including 1319 grade school math test questions.
    \item Code Generation: We utilize HumanEval dataset~\citep{chen2021evaluating} to evaluate the understanding and code-writing capabilities of LLMs. 
    HumanEval contains 164 programming questions to evaluate the language understanding and code-writing capabilities of LLMs, which we refer to as CodeX.
    \item Natural Language Inference: We evaluate this ability of LLMs using two widely utilized datasets: HellaSwag~\citep{zellers2019hellaswag} and TruthfulQA~\citep{lin2021truthfulqa}. 
    HellaSwag contains 70K multiple-choice questions related to commonsense inference.
    TruthfulQA contains 817 questions spanning 38 categories, where the questions are carefully designed to potentially produce incorrect answers due to misconceptions.
    \item Knowledge Reasoning: We assess this ability using the ARC Challenge dataset (ARC-C~\citep{clark2018think}). 
    ARC-C includes 2590 science exam questions from grade 3 to grade 9, which require powerful knowledge and reasoning to complete.
\end{itemize}

For each evaluation dataset, the number of few-shot examples and the evaluation metric used are as shown in Table~\ref{tab:accents}.

\begin{table}
    \centering
    \begin{tabular}{lll}
      \hline
      \textbf{Evalset} & \textbf{Metric}  & \textbf{shots} \\ 
      \hline
      MMLU & acc & 5 \\
      GSM & acc & 5 \\
      HellaSwag & acc-norm & 10 \\
      ARC-C & acc-norm & 25 \\
      TruthfulQA & bleurt-acc & 0 \\
      HumanEval & pass@1 & 0 \\
      \hline
    \end{tabular}
    \caption{Detailed information of our evaluation settings. For each evaluation dataset, we provide the few-shot number and metric used for evaluation.}
    \label{tab:accents}
\end{table}

\subsection{Baselines}
We compare our method with the following baselines:
\begin{itemize}[itemsep=0pt,topsep=0pt]
    \item DEITA~\citep{liu2023makes} proposes a data-efficient instruction tuning approach that automatically evaluates instruction complexity and response quality using LLM-based scorers, and applies diversity-based sampling to select high-quality instruction data.
    \item MoDS~\citep{du2023mods} introduces a model-oriented data selection method that combines quality, coverage, and necessity dimensions to select instruction data by evaluating whether LLMs can correctly respond to given instructions.
    \item IFD~\citep{li2024quantity} develops a self-guided metric for data selection, namely the Instruction-Following Difficulty (IFD) metric.
    \item NUGGETS~\citep{li2023one} constructs a scoring system based on a predefined task set to evaluate whether the data can significantly improve performance across diverse tasks.
    \item ClusterClip~\citep{shao-etal-2024-balanced} uses clustering to reflect the distribution of data and balances the common samples and rare samples.
    \item SelectIT~\citep{liu2024selectit} leverages the intrinsic uncertainty present in LLMs with different parameter sizes to select high-quality data.
    \item InsTag~\citep{lu2023instag} uses ChatGPT to obtain 6.6K tags to comprehensively describe user queries and selects diverse and complex samples based on the tags.
\end{itemize}

\subsection{Experiments Setup}
{
\setlength{\parindent}{0cm}
\textbf{Core Set Selection.}
}
When selecting the core set from the Alpaca-GPT4 dataset using our method, we employ two different LLMs to extract the activation tags: Llama-3.2-3B-Instruct and Llama-3.1-8B-Instruct~\citep{dubey2024llama}.

\begin{table*}[ht]
  \centering
  \begin{tabular}{lccccccc}
    \hline
    \textbf{Method} & \textbf{MMLU} & \textbf{GSM} & \textbf{HellaSwag} & \textbf{TruthfulQA} & \textbf{ARC-C} & \textbf{CodeX} & \textbf{Imp.} \\
    \hline
    \multicolumn{2}{l}{\textit{Base Model: Llama-2-7B}} & & & & & & \\
    Full & 46.40 &	18.65 &	81.25 &	56.06 &	54.18 &	15.61 &	- \\
    DEITA &	47.82 &	16.75 &	\textbf{81.67} &	53.98 &	57.86 &	17.01 &	\cellcolor[rgb]{0.749,0.894,0.992}0.91\% \\
    MoDS &	47.30 &	17.21 &	81.04 &	52.63 &	55.85 &	16.77 &	\cellcolor[rgb]{0.827,0.925,0.992}-0.27\% \\
    IFD &	\textbf{47.84} &	15.31 &	80.82 &	41.62 &	57.19 &	15.85 &	\cellcolor[rgb]{0.941,0.976,0.996}-5.67\% \\
    NUGGETS &	47.70 &	17.13 &	81.32 &	54.71 &	\textbf{58.19} &	17.07 &	\cellcolor[rgb]{0.651,0.851,0.988}1.51\% \\
    ClusterClip &	46.87 &	18.87 &	81.02 &	\textbf{59.73} &	56.52 &	14.82 &	\cellcolor[rgb]{0.714,0.878,0.988}1.29\% \\
    SelectIT & 47.50 &	18.87 &	81.34 &	59.61 &	55.85 &	16.52 &	\cellcolor[rgb]{0.522,0.796,0.984}3.15\%  \\
    InsTag &	46.98 &	18.87 &	81.33 &	57.28 &	55.85 &	15.06 &	\cellcolor[rgb]{0.780,0.925,0.992}0.71\% \\
    $\rm MADS_{3B}$ &	47.46 &	\textbf{19.71} &	81.31 &	57.16 &	56.86 &	\textbf{18.05} &	\cellcolor[rgb]{0.416,0.753,0.980}5.10\% \\
    $\rm MADS_{8B}$ &	47.54 &	19.33 &	81.26 &	60.83 &	57.19 &	17.44 &	\cellcolor[rgb]{0.384,0.737,0.980}5.32\% \\
    \hdashline
    \multicolumn{2}{l}{\textit{Base Model: Llama-3-8B}} & & & & & & \\
    Full &	63.59 &	57.92 &	84.16 &	65.24 &	\textbf{63.55} &	46.28 &	- \\
    DEITA &	63.65 &	58.76 &	84.45 &	58.75 &	59.52 &	45.98 &	\cellcolor[rgb]{0.749,0.894,0.992}-2.51\% \\
    MoDS &	64.48 &	61.25 &	84.43 &	58.07 &	61.54 &	45.00 &	\cellcolor[rgb]{0.827,0.925,0.992}-1.57\% \\
    IFD &	64.71 &	58.53 &	83.38 &	50.43 &	60.87 &	42.99 &	\cellcolor[rgb]{0.941,0.976,0.996}-5.36\% \\
    NUGGETS &	64.72 &	60.50 &	84.11 &	61.08 &	61.20 &	43.90 &	\cellcolor[rgb]{0.886,0.953,0.996}-1.51\% \\
    ClusterClip &	64.88 &	60.20 &	83.94 &	64.38 &	61.87 &	47.86 &	\cellcolor[rgb]{0.780,0.925,0.992}0.86\% \\
    SelectIT &	\textbf{64.91} &	59.74 &	84.02 &	64.50 &	60.54 &	45.43 &	\cellcolor[rgb]{0.827,0.925,0.992}-0.44\% \\
    InsTag &	64.82 &	60.65 &	\textbf{84.28} &	63.04 &	62.21 &	48.35 &	\cellcolor[rgb]{0.749,0.894,0.992}0.96\% \\
    $\rm MADS_{3B}$ &	64.89 &	59.89 &	84.24 &	65.24 &	61.87 &	\textbf{48.96} &	\cellcolor[rgb]{0.651,0.851,0.988}1.45\% \\
    $\rm MADS_{8B}$ &	64.86 &	\textbf{60.72} &	84.07 &	\textbf{68.30} &	62.54 &	48.23 &	\cellcolor[rgb]{0.553,0.808,0.984}2.34\% \\
    \hline
  \end{tabular}
  \caption{The overall results on the benchmark tasks.
  ``Full'' denotes training with the complete dataset, while all other methods use 15\% of the full dataset for training.
  $\rm MADS_{3B}$ denotes core set selection using Llama-3.2-3B-Instruct, while $\rm MADS_{8B}$ denotes core set selection using Llama-3.1-8B-Instruct.
  "Imp." denotes the average improvement across all tasks.}
  \label{tab:main}
\end{table*}

\begin{table*}[!ht]
  \centering
  \begin{tabular}{lccccccc}
    \hline
    \textbf{Layer Selection} & \textbf{MMLU} & \textbf{GSM} & \textbf{HellaSwag} & \textbf{TruthfulQA} & \textbf{ARC-C} & \textbf{CodeX} & \textbf{Imp.} \\
    \hline
    2 & 46.22 & 15.62 & 81.44 & 58.63 & 56.86 & 16.03 & -0.70\% \\
    
    9 & 46.98 & 16.98 & \textbf{81.56} & 56.18 & 57.86 & \textbf{17.74} & 2.22\% \\
    16 & 46.60 & 15.62 & 81.30 & \textbf{59.97} & \textbf{58.86} & 16.46 & 0.88\% \\
    21 & 46.96 & \textbf{17.89} & 81.22 & 57.41 & 57.86 & 16.16 & 1.64\% \\
    28 & \textbf{47.1} & 17.51 & 81.15 & 57.89 & 58.19 & 17.07 & 2.55\% \\
    \hdashline
    2-9 & 46.21 & 15.84 & 81.01 & 56.55 & \textbf{57.86} & 17.62 & 0.80\% \\
    2-9-16 & 46.6 & 18.04 & 81.30 & 59.00 & 56.52 & \textbf{19.15} & 4.91\% \\
    2-9-16-21 & 47.1 & 17.44 & 81.06 & \textbf{59.36} & 56.86 & 17.07 & 2.50\% \\
    2-9-16-21-28 & \textbf{47.46} & \textbf{19.71} & \textbf{81.31} & 57.16 & 56.86 & 18.05 & \textbf{5.10\%} \\
    All Layers & 46.05 & 18.65 & 81.27 & 59.98 & 57.19 & 17.62 & 4.12\% \\
    \hdashline
    2-9-16-28 & 46.25 & 18.50 & 81.35 & 58.87 & 57.53 & 16.28 & 2.41\% \\
    \hdashline
    7 & 46.95 & 16.30 & 81.39 & 60.59 & 57.19 & 15.91 & 0.72\% \\
    2-7-16-21-28 & 47.50 & 17.97 & 80.90 & 59.36 & 55.85 & 17.01 & 2.71\% \\
    2-7-9-16-21-28 & 45.16 & 16.68 & 80.83 & 59.73 & 58.53 & 17.07 & 1.70\% \\
    \hline
  \end{tabular}
  \caption{Performance on downstream tasks of Llama-2-7B fine-tuned with core sets selected by activation tags from each layer chosen through our strategy. 
    "All Layers" denotes using the activation tags from all 28 layers of Llama-3.2-3B-Instruct for core set selection.}
  \label{tab:layer}
  \vspace{-0.2cm} 
\end{table*}

{
\setlength{\parindent}{0cm}
\textbf{Training Details.}
}
We fine-tune the Llama-2-7B~\citep{touvron2023llama} base model and Llama-3-8B~\citep{dubey2024llama} base model. 
For the Llama-2-7B base model, we fine-tune it for 3 epochs, with a batch size of 128, using the Adam optimizer with a $1 \times 10^{-5}$ learning rate and a 0.01 warm-up ratio.
For the Llama-3-8B base model, we fine-tune it for 6 epochs, with a batch size of 256, using the Adam optimizer with a $2 \times 10^{-6}$ learning rate and a 0.06 warm-up ratio.

\subsection{Preliminary Validation of Activation-Tag Representations}
\label{sec:preliminary_validation}
\begin{figure*}[t]
    \centering
    \includegraphics[width=0.32\textwidth]{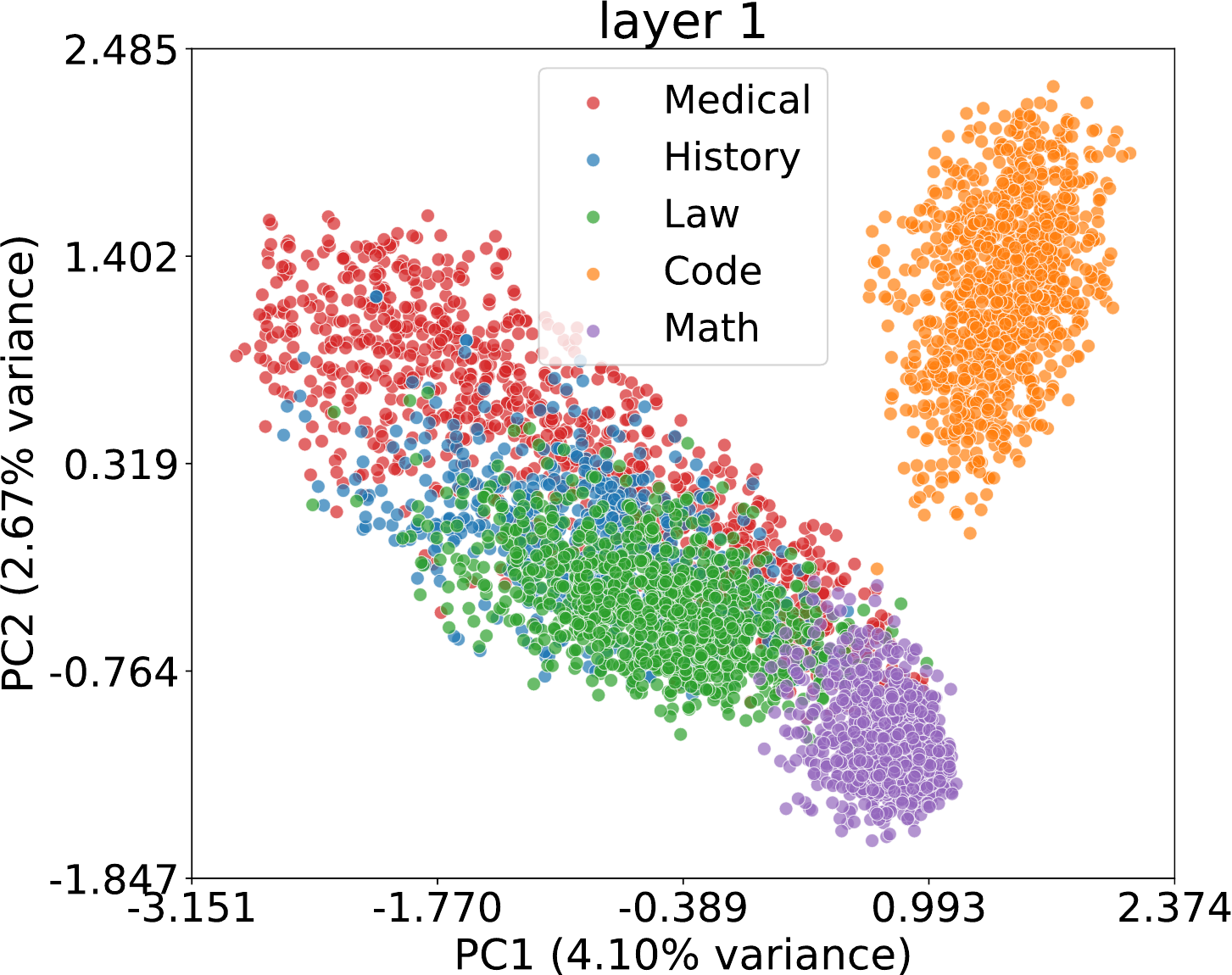}
    \includegraphics[width=0.32\textwidth]{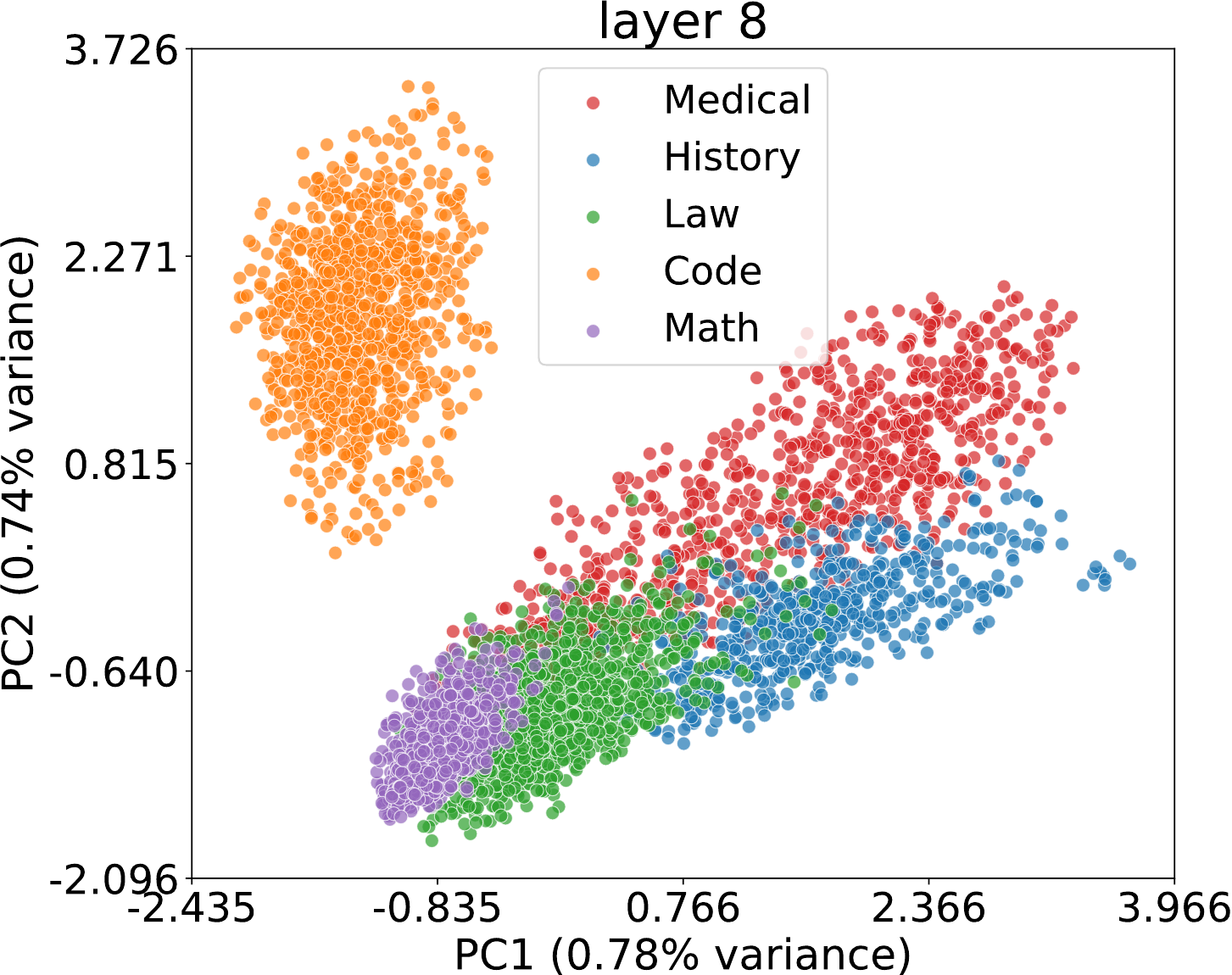}
    \includegraphics[width=0.32\textwidth]{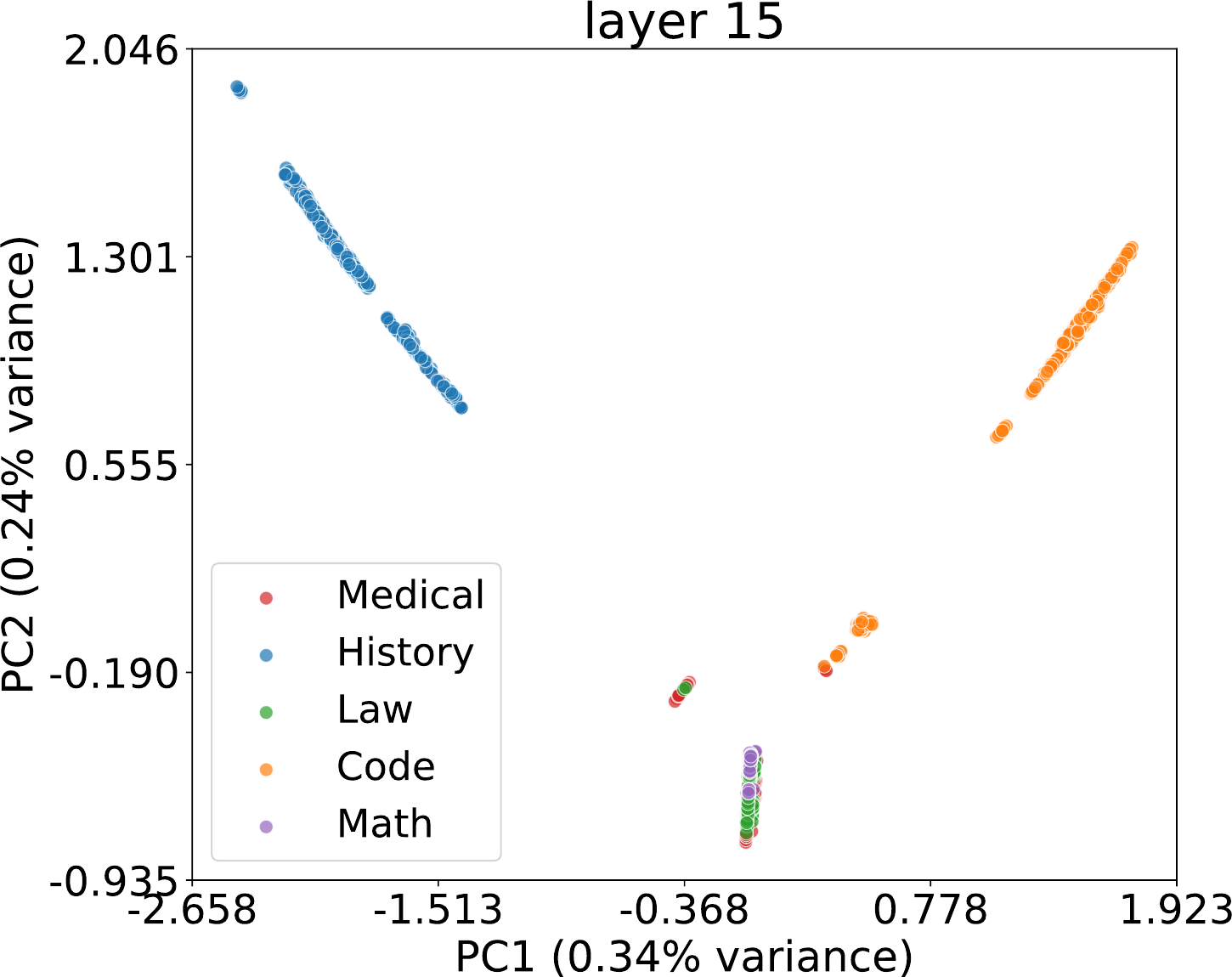}
    \caption{
        PCA visualization of activation tag vectors at layer 1, 8 and 15 of Llama-3.2-3B-Instruct for five instruction categories: code (orange), math (purple), law (green), medical (red), and history (blue). 
        Instructions are encoded as $c$-dimensional one-hot vectors, where $c$ is the total number of distinct activation tags, and reduced to 2D using PCA.
        The clustering demonstrates that activation tags capture domain-specific semantic features.
    }
    \label{fig:pca_visualization}
\end{figure*}

\begin{figure*}[t]
    \centering
    \includegraphics[width=0.30\textwidth]{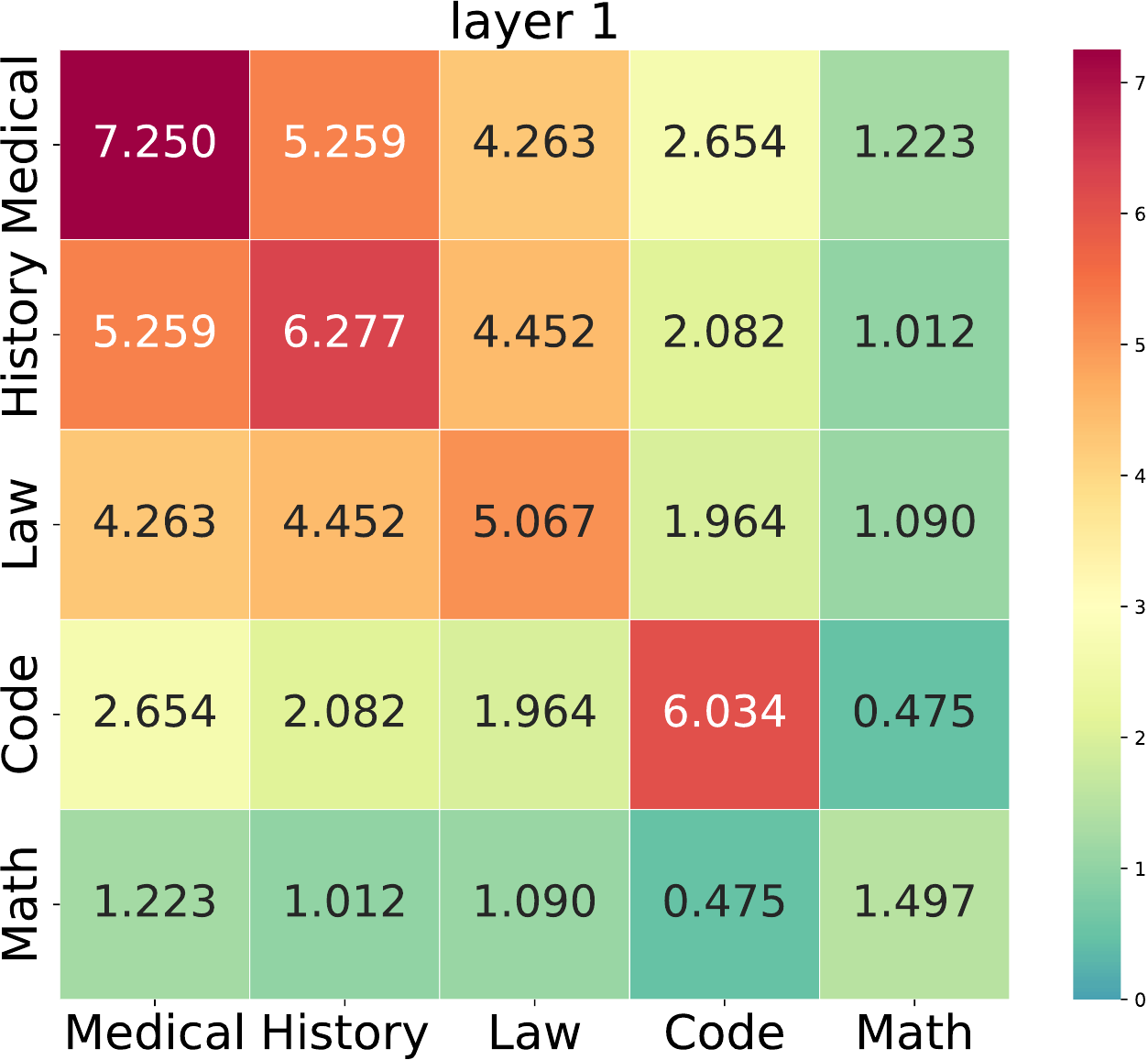}
    \includegraphics[width=0.30\textwidth]{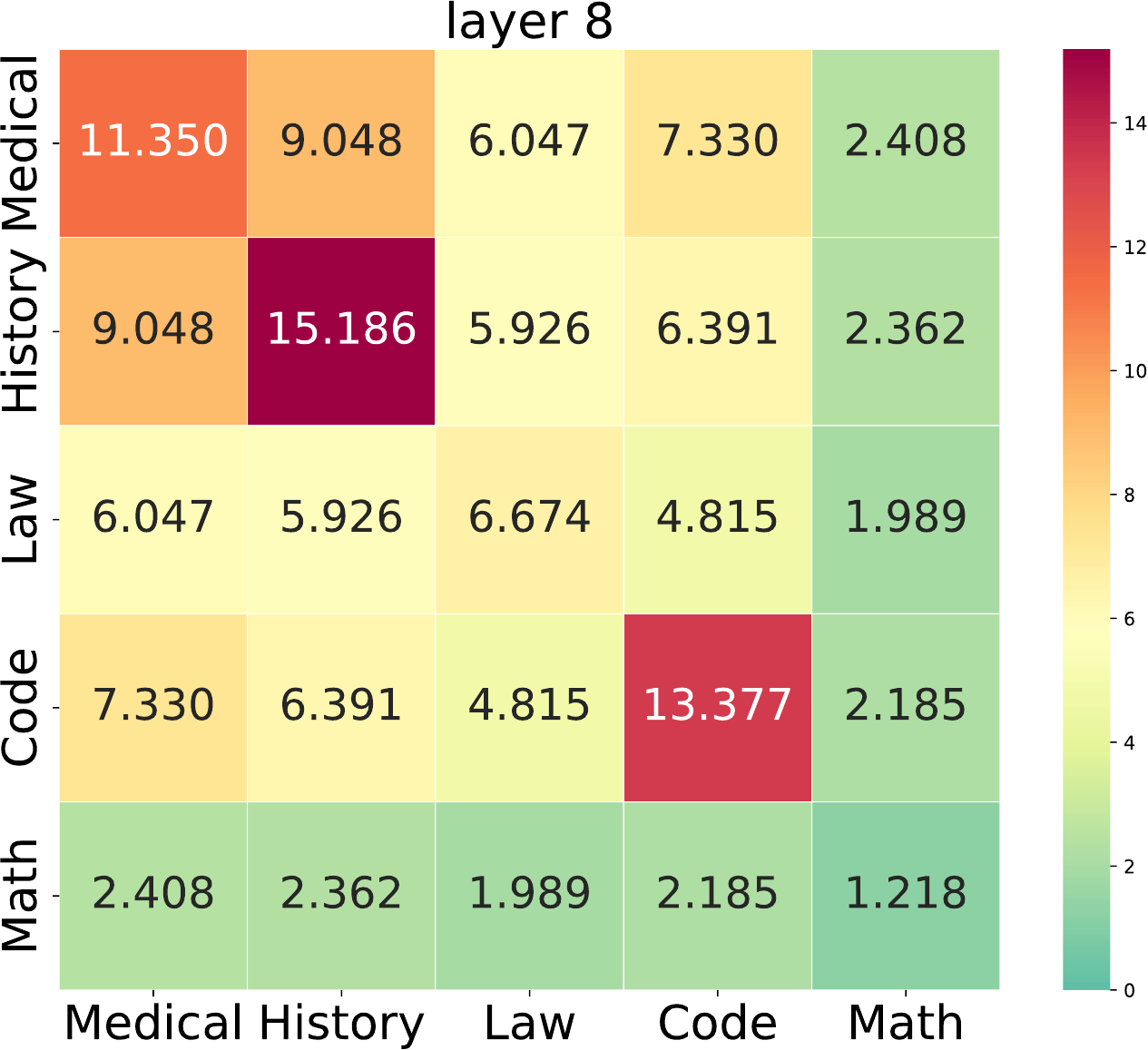}
    \includegraphics[width=0.30\textwidth]{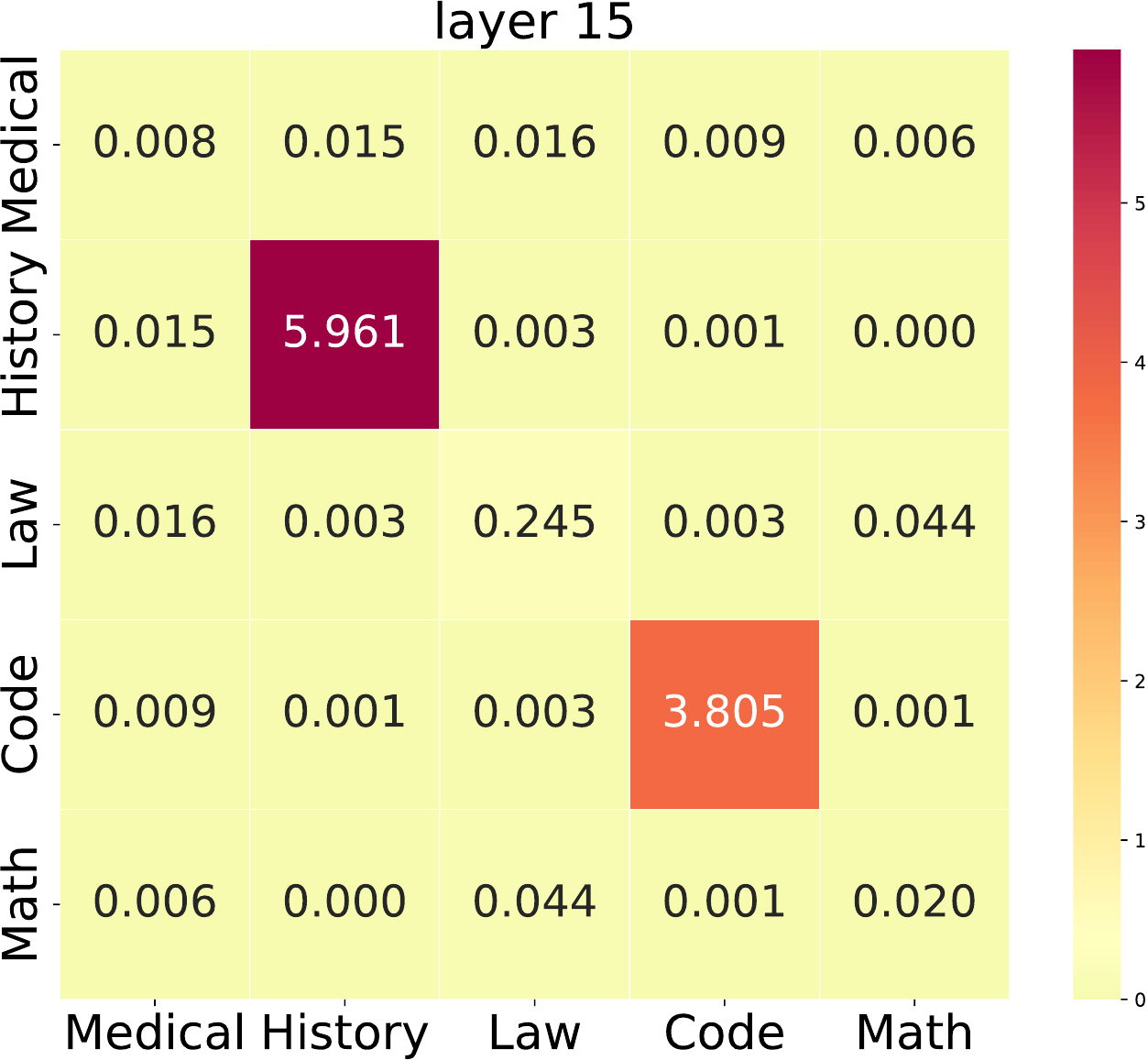}
    \caption{
      Heatmap of the average number of shared activation tags between instruction categories at layers 1, 8, and 15 of Llama-3.2-3B-Instruct.
      Each cell $(i, j)$ represents the average number of activation tags shared between instruction pairs from domain $i$ and domain $j$.
      Most diagonal values (intra-domain similarity) are higher than off-diagonal values (cross-domain similarity), with darker colors indicating more shared tags.
      This pattern confirms that same-domain instructions share more tags than different-domain pairs, validating the correlation between activation patterns and data features.
    }
    \label{fig:similarity_heatmap}
\end{figure*}

In this subsection, we empirically validate that neuron activation tags capture domain-specific semantic features, providing the foundational motivation for MADS.
We conduct experiments using 1,000 randomly selected instructions from each of five domains: code\footnote{\href{https://huggingface.co/datasets/ise-uiuc/Magicoder-OSS-Instruct-75K}{https://huggingface.co/datasets/ise-uiuc/Magicoder-OSS-Instruct-75K}}, mathematics\footnote{\href{https://huggingface.co/datasets/openai/gsm8k}{https://huggingface.co/datasets/openai/gsm8k}}, legal analysis\footnote{\href{https://huggingface.co/datasets/cais/mmlu/tree/main/professional_law}{https://huggingface.co/datasets/cais/mmlu/tree/main/professional_law}}, medical consultation\footnote{\href{https://huggingface.co/datasets/hongzhouyu/FineMed-SFT}{https://huggingface.co/datasets/hongzhouyu/FineMed-SFT}}, and historical Q\&A\footnote{\href{https://huggingface.co/datasets/nielsprovos/world-history-1500-qa}{https://huggingface.co/datasets/nielsprovos/world-history-1500-qa}}. 
We extract activation tags at five layers of Llama-3.2-3B-Instruct, and perform two analyses:
(1) \textbf{PCA Visualization}: For each instruction, we construct a $c$-dimensional one-hot vector, where $c$ is the total number of distinct activation tags across all five domains. 
Each dimension corresponds to an activation tag, set to 1 if present in the instruction and 0 otherwise. 
We apply PCA to reduce these vectors to 2D for visualization, revealing domain-specific clustering (Figure~\ref{fig:pca_visualization}).
(2) \textbf{Similarity Analysis}: We compute the average number of shared activation tags between instruction pairs, comparing intra-domain pairs with cross-domain pairs. 
Results show that same-domain instructions share significantly more tags than different-domain pairs (Figure~\ref{fig:similarity_heatmap}).
For brevity, we display the results for three representative layers here; the results for the other two layers are provided in Appendix~\ref{appendix:visualization}.
Both experimental results indicate a correlation between activation tags and data features.
Moreover, both experiments reveal a consistent pattern: code and historical Q\&A instructions exhibit the largest separation from other domains. 
While there are distinct differences in activation tags among mathematics, legal, and medical instructions, the variances between these three categories are relatively smaller compared to code and historical instructions. 
This aligns with intuition: code instructions contain syntactically unique programming languages, and historical Q\&A focuses on factual knowledge retrieval, whereas the other three domains require logical reasoning to derive answers.

To provide a more grounded interpretation beyond domain labels alone, we further examine the dataset-internal characteristics of the sampled instructions. (see Appendix~\ref{appendix:domain_examples} for representative examples). 
For the code domain, 60.4\% of instructions contain embedded code blocks (marked by backtick delimiters), and 74.3\% include programming-specific tokens (\texttt{class}, \texttt{def}, \texttt{import}, \texttt{return}, etc.) that are virtually absent in other domains—this lexically unique structure produces highly distinctive activation patterns. 
For the historical Q\&A domain, we note that although individual historical questions may indeed involve multi-factor reasoning (e.g., analyzing political, economic, or cultural dimensions of historical events), all sampled instructions share a uniform document-comprehension format: each presents a primary source passage and asks the model to select the correct interpretation from multiple-choice options, without requiring the model to retrieve or generate knowledge from memory. 
This highly consistent task structure—not merely domain-level content—likely contributes substantially to the distinct activation clustering observed. 
By contrast, mathematical, legal, and medical instructions exhibit greater internal heterogeneity in task format and reasoning demands: 
mathematical instructions predominantly require procedural step-by-step arithmetic reasoning (98.6\% following explicit chain-of-thought computation notation); 
legal instructions span constitutional analysis (16.7\%), criminal law scenarios (23.8\%), and civil procedure cases (11.8\%), encompassing both rule-retrieval and multi-step legal reasoning; 
and medical instructions mix symptom-based clinical consultations (40.6\%) and open-ended knowledge explanation (26.6\%). 
This greater task diversity within these three domains, combined with their shared reliance on natural language analytical reasoning, leads to more overlapping activation patterns and explains the relatively smaller inter-domain separation observed among them.

\section{Results}
\subsection{Main Results}

We applied our method separately to neuron activation data from Llama-3.2-3B-Instruct and Llama-3.1-8B-Instruct, using a core set size of 15\% of the full dataset.
The experimental results are presented in Table~\ref{tab:main}.
From the results, we derive the following observations:

Our method significantly outperforms the full dataset and other baselines in terms of average improvement across all tasks.
While some methods excel in specific tasks, they underperform in others, indicating that our method enhances LLM capabilities without sacrificing specific abilities.
This is attributed to our method's focus on ensuring diversity and coverage within the core set.

Our method shows robustness in core set selection across different models.
Comparing the core set selection results of Llama-3.2-3B-Instruct and Llama-3.1-8B-Instruct, we find that our method effectively performs the core set selection task and both achieve the best results, regardless of whether the model has a similar parameter size to the base model or a smaller parameter size.
This indicates that our method provides an effective approach to leverage smaller LLMs to enhance the capabilities of larger LLMs, which is important for selecting datasets under resource constraints.

\subsection{Analysis}

We conducted detailed analysis experiments to answer the following questions:
(1) How does the layer selection strategy affect our method's effectiveness? (\S \ref{sec:layer_selection_strategy}) 
(2) How do different selected layers and their interactions influence task-specific improvements? (\S \ref{sec:layer}) 
(3) How does filtering low-frequency activation tags affect downstream tasks? (\S \ref{sec:frequency}) 
(4) How robust is MADS across different base models? (\S \ref{sec:robust}) 
(5) How does activation-based coverage differ from embedding-based coverage in terms of redundancy and feature coverage? (\S \ref{sec:redundancy})
(6) How does the rounding granularity of $v_{tag}$ affect the grouping mechanism and downstream task performance? (\S \ref{sec:vtag_analysis})

\subsubsection{Effect of Layer Selection Strategy}
\label{sec:layer_selection_strategy}
\begin{figure}[!t]
    \centering
    \includegraphics[width=0.98\columnwidth]{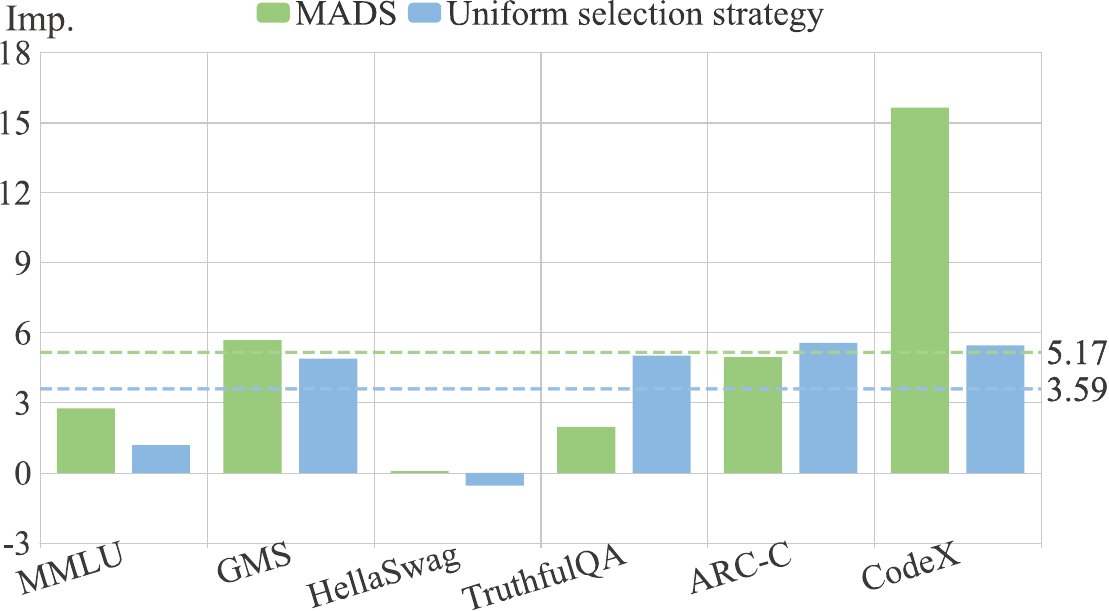}
    \caption{Comparison of two layer selection strategies on Llama-2-7B fine-tuning performance.
    The x-axis shows six benchmark tasks, and the y-axis shows relative improvement over the full dataset baseline.
    Green bars represent our peak-based strategy (selecting layers 2, 9, 16, 21, 28), achieving 5.17\% average improvement.
    Blue bars represent the uniform selection strategy (selecting layers 1, 8, 15, 22, 28), achieving 3.59\% average improvement.
    Our strategy outperforms uniform selection by 1.58\% on average.}
    \label{fig:layer_selection_strategy}
\end{figure}

In our method, during the activation tag filtering stage, we select $M$ layers where the proportion of strongly activated neurons is locally maximal.
To verify this strategy's effectiveness, we compared it with a uniform selection of $M$ layers, which is a more direct and straightforward selection approach.
Specifically, when using the activation tags of Llama-3.2-3B-Instruct, which consists of 28 layers, our method selects the 2nd, 9th, 16th, 21st, and 28th layers.
The uniform selection strategy selects the 1st, 8th, 15th, 22nd, and 28th layers.
We fine-tuned Llama-2-7B with core sets from both strategies.
Results in Figure~\ref{fig:layer_selection_strategy} show that both strategies can improve the model's capabilities, but our strategy results in an average improvement of 5.17\%, compared to 3.59\% with the uniform strategy.

\subsubsection{Effect of Different Layers}
\label{sec:layer}
The activation tags used for core set selection are sampled from different layers.
To validate (1) the impact of the number of selected layers on LLM capabilities and (2) the effect of different layers' activation tags, we applied our method using activation tags from different layers and then fine-tuned Llama-2-7B.

From the results in Table~\ref{tab:layer}, we observed that as the number of sampled layers increased, the overall capability of the LLM generally improved, and the best results were obtained when using the 2nd, 9th, 16th, 21st, and 28th layers.
An exception was found with the 2nd, 9th, 16th, and 21st layers.
To verify whether the 21st layer has a negative impact on the final results, we further used the 2nd, 9th, 16th, and 28th layers, and found that the results were significantly worse than when using these five layers.
We hypothesize that the 21st layer can work in conjunction with the 28th layer to exert its effect, suggesting potential synergistic interactions between layers.

In order to further study whether the activation tags of different layers have different effects on improving LLM capabilities, we performed core set selection based on the activation tags of individual layers.
From the results in Table~\ref{tab:layer}, it can be observed that different layers exhibit distinct enhancement effects for different tasks.
For instance, the 16th layer showed significant improvements in TruthfulQA and ARC-C tasks, but performed poorly in other tasks, such as MMLU and GSM.
Our method's multi-layer core set selection allows mutual compensation, achieving balanced LLM enhancement.
Furthermore, we conducted an experiment using all layers for core set selection. 
As shown in Table~\ref{tab:layer}, utilizing all layers achieves a 4.12\% average improvement, which is still lower than the 5.10\% improvement achieved by our layer selection strategy.
This result confirms that while incorporating more layers provides more information, our selective approach effectively identifies the most informative layers and avoids redundant information.

To validate our layer selection strategy for handling adjacent peaks, we conducted additional experiments on layer 7, which is a local maximum but was excluded from our final selection due to its proximity to layer 9 (separated by only one layer). As shown in Table~\ref{tab:layer}, when using layer 7 alone for core set selection, the average improvement is only 0.72\%, which is significantly lower than layer 9's improvement of 2.22\%. 
Additionally, when including layer 7 in the multi-layer selection (2-7-16-21-28), the average improvement drops to 2.71\%, compared to 5.10\% when using layer 9 instead. 
Furthermore, to directly examine the impact of including both adjacent peaks simultaneously, we conducted an experiment using layers 2-7-9-16-21-28. 
As shown in Table~\ref{tab:layer}, although using both layer 7 and layer 9 achieves strong performance on TruthfulQA and ARC-C tasks, it exhibits notable degradation on MMLU and GSM tasks.
The results indicate that violating the even distribution principle by selecting adjacent peaks impedes balanced enhancement across multiple capabilities, leading to uneven improvements on different tasks.
This experiment provides direct empirical evidence that excluding adjacent peaks is essential for optimal performance.

These results are consistent with prior findings on layerwise processing in Transformers: representations are hierarchically organized (earlier layers are more lexical/syntactic and middle-to-late layers become increasingly semantic/discourse or prediction-related)~\citep{li2025echoesbert,he2024decodingprobing}, while adjacent layers tend to be more representationally similar than distant ones~\citep{jiang2024layerwisesimilarity}. 
Therefore, our choice of (2,9,16,21,28) can be viewed as a lightweight way to sample heterogeneous feature regimes across layers and reduce redundancy from neighboring layers.

\subsubsection{Effect of Activation Tags Filtering}
\label{sec:frequency}

\begin{figure}[!t]
    \centering
    \includegraphics[width=0.98\columnwidth]{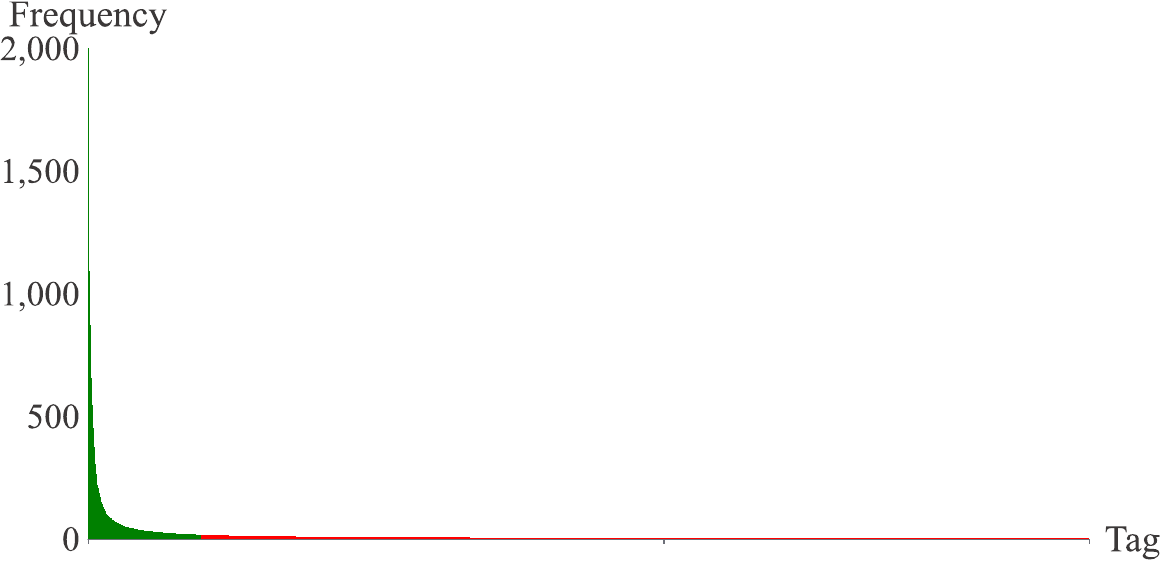}
    \caption{Long-tail distribution of activation tag frequencies at layer 15 of Llama-3.2-3B-Instruct.
    The x-axis represents activation tags sorted by frequency in descending order, and the y-axis represents the frequency (number of instructions containing each tag).
    Green region indicates high-frequency tags retained in the core set, while red region indicates low-frequency tags filtered out as noise.
    This filtering strategy removes rare activation patterns that lack generality while preserving representative activation patterns.}
    \label{fig:fre}
\end{figure}

To demonstrate the rationality of filtering out low-frequency activation tags in our method, we conducted a statistical analysis of the frequency $f(tag)$ of each activation tag.
We found that they present a long-tail distribution, where a few activation tags appear frequently, while most appear infrequently.
Our method filters out low-frequency activation tags, retaining only those at the head of the long-tail distribution, as illustrated in Figure~\ref{fig:fre}.

\begin{figure}[!t]
    \centering
    \includegraphics[width=0.95\columnwidth]{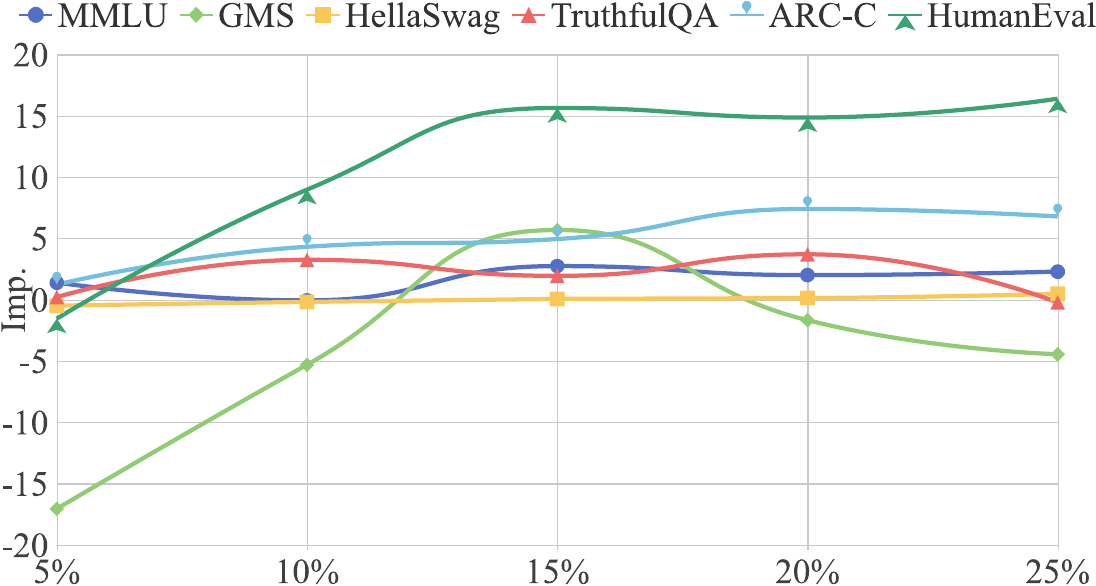}
    \caption{Performance of Llama-2-7B on six benchmark tasks with varying core set sizes.
    The x-axis represents the percentage of Alpaca-GPT4 data used (5\%--25\%), controlled by threshold $\theta_{base}$ (58, 28, 17, 12, 10 respectively).
    The y-axis shows accuracy on each task.
    The optimal data size varies by task, with 15\% achieving the best average improvement.}
    \label{fig:data_size_line}
\end{figure}

We further explored the impact of activation tags filtering on the performance of LLMs on different tasks.
In our method, by adjusting the frequency filtering threshold $\theta_{base}$, we can control the number of filtered activation tags, thus controlling the size of the core set.
When using Llama-3.2-3B-Instruct for core set selection, we set the size of the core set to 5\%, 10\%, 15\%, 20\%, and 25\% of the original dataset, with corresponding $\theta_{base}$ values of 58, 28, 17, 12 and 10.
We then fine-tuned Llama-2-7B using these core sets and evaluated the performance on different tasks, as shown in Figure~\ref{fig:data_size_line}.
The experiments show that the optimal $\theta_{base}$ varies for different tasks; for example, GSM achieved the best results when $\theta_{base} = 17$ (15\% of the data), while ARC-C achieved the best results when $\theta_{base} = 12$ (20\% of the data).
Moreover, the sensitivity to $\theta_{base}$ varies across different tasks.
For instance, the performance on the HellaSwag task remains relatively stable regardless of changes in $\theta_{base}$, while the performance on the GSM task exhibits significant fluctuations with variations in $\theta_{base}$.

\begin{table*}
  \centering
  \begin{tabular}{lccccccc}
    \hline
    \textbf{Method} & \textbf{MMLU} & \textbf{GSM} & \textbf{HellaSwag} & \textbf{TruthfulQA} & \textbf{ARC-C} & \textbf{CodeX} & \textbf{Imp.} \\
    \hline
    \multicolumn{2}{l}{\textbf{\textit{Base Model: Llama-2-7B}}} & & & & & & \\
    Full & 48.23 & 19.78 & 80.19 & 52.14 & 53.85 & 19.02 &	- \\
    $\rm MADS_{3B}(10\%)$ & 47.75 & 20.47 & 79.24 & 53.37 & 54.85 & 18.84 & 0.76\% \\
    $\rm MADS_{3B}(15\%)$ & 47.75 & 19.33 & 79.47 & 51.90 & 54.18 & 20.12 & 0.29\% \\
    \hline
  \end{tabular}
  \caption{Performance of MADS for core set selection on the WizardLM Dataset} 
  \label{table:robustness_dataset}
\end{table*}

\subsubsection{Robustness across Models and Datasets}
\label{sec:robust}
\begin{figure}[!t]
    \centering
    \includegraphics[width=0.98\columnwidth]{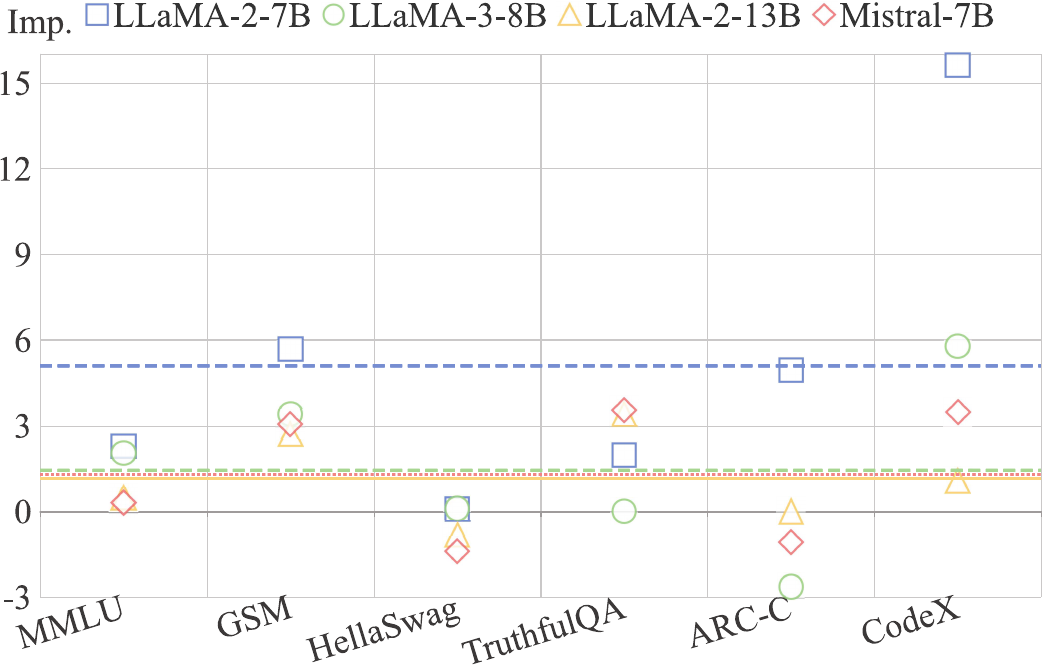}
    \caption{Performance improvement of different base models fine-tuned with the core set selected by Llama-3.2-3B-Instruct compared to the full dataset.
    The x-axis shows four base models: Llama-2-7B, Llama-2-13B, Llama-3-8B, and Mistral-7B.
    Bars represent improvement on individual tasks, and horizontal lines indicate average improvement across all six benchmark tasks.
    All models show positive average improvement, demonstrating that the core set selected by Llama-3.2-3B-Instruct generalizes effectively across different model families and scales.}
    \label{fig:diff_model}
\end{figure}

To verify the robustness of MADS, we tested its performance on various base models.
We used the full Alpaca-GPT4 dataset and the core set containing 15\% of the data selected based on Llama-3.2-3B-Instruct to fine-tune different base models, as shown in Figure~\ref{fig:diff_model}.
The experimental results demonstrate that, in terms of average performance improvement across all benchmark tasks, MADS enhances the capabilities of both Llama 2 and Llama 3 models, regardless of their parameter sizes.
Additionally, although the core set was selected by Llama-3.2-3B-Instruct, it was also applicable to Mistral-7B~\citep{jiang2023mistral}.
This experiment demonstrates the generalizability of MADS which does not rely on a specific base model but can effectively improve the capabilities of LLMs of different series or scales.

Furthermore, to validate the robustness of our method across different datasets, we performed core set selection on the WizardLM dataset based on Llama-3.2-3B-Instruct and subsequently fine-tuned Llama-2-7B. 
As shown in Table~\ref{table:robustness_dataset}, we experimented with both 10\% and 15\% of the WizardLM data. The results demonstrate that both settings achieve performance comparable to that obtained using the full dataset, which validates the robustness of our method. 
Notably, using 10\% of the data yields the best improvement, which is different from the results on the Alpaca-GPT4 dataset where 15\% of the data performed best.
This discrepancy may be attributed to the larger size of the WizardLM dataset compared to the Alpaca-GPT4 dataset, which provides greater potential for compression when selecting diverse data, in addition to differences in the inherent quality of the datasets.

\subsubsection{Case Study}
Our method assumes different neuron activations correspond to different data features. To verify this, we present data cases corresponding to different activation tags in Table~\ref{tab:cases}.
Data with identical activation labels exhibit similar features, validating the rationale of our method.

\begin{table*}[ht]
    \centering
    \begin{tabular}{ll}
      \hline
      \multicolumn{2}{l}{\textit{layer 1}} \\
      \hdashline
      \multicolumn{2}{l}{\textit{7045: Activated when a specific quantity is present in the instruction}} \\
      $\mathrm{[Instruction]}$ & Give \textcolor{orange}{three} tips for staying healthy. \\
      $\mathrm{[Instruction]}$ & List \textcolor{orange}{five} factors that lead to global warming. \\
      $\mathrm{[Instruction]}$ & Sort the following items into \textcolor{orange}{two} categories. \\
      \hdashline
      \multicolumn{2}{l}{\textit{4052: Activated when a sequence occurs in an instruction}} \\
      $\mathrm{[Instruction]}$ & Arrange the given numbers in ascending order. \textcolor{orange}{2, 4, 0, 8, 3} \\
      $\mathrm{[Instruction]}$ & Convert the given binary number to its decimal equivalent. \textcolor{orange}{101101} \\
      $\mathrm{[Instruction]}$ & Generate a sentence that contains the given words. \textcolor{orange}{magic, castle, king} \\
      \hline
      \hline
      \multicolumn{2}{l}{\textit{layer 8}} \\
      \hdashline
      \multicolumn{2}{l}{\textit{4990: Activated when math is present in the instruction}} \\
      $\mathrm{[Instruction]}$ & Explain why the following fraction is equivalent to \textcolor{orange}{1/4}: \textcolor{orange}{4/16}  \\
      $\mathrm{[Instruction]}$ & Variable x is defined as “\textcolor{orange}{4x + 2y = 10}”. Find the value of x. \\
      $\mathrm{[Instruction]}$ & What is the force on a \textcolor{orange}{1 kg} mass due to the gravitational force? \\
      \hdashline
      \multicolumn{2}{l}{\textit{6082,5712: Activated when the response is required to be a sequence}} \\
      $\mathrm{[Instruction]}$ & Generate \textcolor{orange}{a list of} ten items a person might need for a camping trip  \\
      $\mathrm{[Instruction]}$ & Generate a password of \textcolor{orange}{8 characters}. \\
      $\mathrm{[Instruction]}$ & Generate a sentence using the following words \textcolor{orange}{in the correct order}: house on fire \\
      \hline
      \hline
      \multicolumn{2}{l}{\textit{layer 15}} \\
      \hdashline
      \multicolumn{2}{l}{\textit{8042,596,3602,4562,7275,1557: Activated when a command asks to describe something}} \\
      $\mathrm{[Instruction]}$ & \textcolor{orange}{Describe} the structure of an atom. \\
      $\mathrm{[Instruction]}$ & \textcolor{orange}{Describe} a time when you had to make a difficult decision. \\
      $\mathrm{[Instruction]}$ & \textcolor{orange}{Describe} the function of a computer motherboard.  \\
      \hdashline
      \multicolumn{2}{l}{\textit{596,8042,5358,6963,7042: Activated when a command asks to write something}} \\
      $\mathrm{[Instruction]}$ & \textcolor{orange}{Write} a short paragraph about the given topic. \\
      $\mathrm{[Instruction]}$ & \textcolor{orange}{Write} a review of a recent movie you watched. \\
      $\mathrm{[Instruction]}$ & \textcolor{orange}{Write} a simple definition of the word "economics"  \\
      \hline
    \end{tabular}
    \caption{Examples of activation tags and corresponding instructions.}
    \label{tab:cases}
\end{table*}

\renewcommand{\arraystretch}{0.9}
\begin{table*}
    \centering
    \begin{tabular}{p{0.05\textwidth}p{0.05\textwidth}p{0.9\textwidth}}
      \hline
      \textbf{Tag} & \textbf{$v_{tag}$} & \textbf{Instruction Examples} \\
      \hline
      \multicolumn{3}{l}{\textit{layer1-4665: Activated when computing the median}} \\
      \hdashline
      & 1.0 & \textit{(Median computation on sequence data)} \\
      & & Use the given data to calculate the \textcolor{orange}{median}. \textcolor{orange}{[2, 3, 7, 8, 10]} \\
      & & How do you calculate the \textcolor{orange}{median} from the given data? \textcolor{orange}{1, 2, 8, 9, 12, 13} \\
      & & Create a Python program to calculate the \textcolor{orange}{median} of an array of numbers. \textcolor{orange}{5, 15, 20, 2, 10} \\
      \cdashline{2-3}
      & 1.1 & \textit{(Median computation on non-sequence data)} \\
      & & Given a table of data, calculate the mean, \textcolor{orange}{median}, and range of the data. \\
      & & For the given input, ... Calculating the \textcolor{orange}{median} cost of gas \\
      & & Predict the \textcolor{orange}{median} age of London. \\
      \hline
      \multicolumn{3}{l}{\textit{layer1-7955: Activated when the instruction requires the text to be reorganized.}} \\
      \hdashline
      & 1.0 & \textit{(Re-arrange tasks)} \\
      & & \textcolor{orange}{Re-arrange} the following letters to form a meaningful word. vhics \\
      & & \textcolor{orange}{Re-arrange} this sentence: ``Happy are those who dream'' \\
      & & \textcolor{orange}{Re-arrange} the given words to make it into a valid sentence. the students best performed \\
      \cdashline{2-3}
      & 1.1 & \textit{(Rewrite tasks)} \\
      & & \textcolor{orange}{Rewrite} the following sentence in the third person. I am anxious \\
      & & \textcolor{orange}{Rewrite} the sentence with more descriptive words. The game is fun. \\
      & & \textcolor{orange}{Rewrite} the given sentence using a different but similar word. She partook in the event. \\
      \hline
      \multicolumn{3}{l}{\textit{layer8-819,4424,7062,4990: Activated when multiplicative operations are present}} \\
      \hdashline
      & 1.5 & \textit{(The sentence pattern is "Calculate the product...")} \\
      & & \textcolor{orange}{Calculate the product} of 5 and 3. \\
      & & \textcolor{orange}{Calculate the product} given two numbers. 4 and 8 \\
      & & \textcolor{orange}{Calculate the product} of the two values. 3 and 5 \\
      \cdashline{2-3}
      & 1.2 & \textit{(The sentence pattern is "Find the product of...")} \\
      & & \textcolor{orange}{Find the product} of the numbers. 5 and 8 \\
      & & \textcolor{orange}{Find the product} of 29 and 32 \\
      & & \textcolor{orange}{Find the product} of 1.8 and 5. \\
      \hline
      \multicolumn{3}{l}{\textit{layer8-5712,4567,465: Activated when append actions or addition operations are present}} \\
      \hdashline
      & 1.3 & \textit{(The added quantity is 3)} \\
      & & \textcolor{orange}{Add 3} words to make the sentence more vivid.The teacher gave a speech. \\
      & & \textcolor{orange}{Add 3} more animals to the following list. Dogs, Cats, Monkeys \\
      & & \textcolor{orange}{Add 3} new ingredients to a pasta dish \\
      \cdashline{2-3}
      & 1.5 & \textit{(The added quantity is 5)} \\
      & & \textcolor{orange}{Add 5} items to a grocery shopping list. \\
      & & \textcolor{orange}{Add 5} eights to the number 9. \\
      & & \textcolor{orange}{Add 5px} to each of the current margin values. margin-left: 20px; margin-top: 30px; \\
      \hline
      \multicolumn{3}{l}{\textit{layer8-3221,6071,6082,2225: Activated when the instruction requires describing something.}} \\
      \hdashline
      & 1.8 & \textit{(Describe the 'scenario.')} \\
      & & \textcolor{orange}{Describe a scenario} where someone could be accused of plagiarism. \\
      & & \textcolor{orange}{Describe a scenario} where a student's choices can either cause success or failure. \\
      & & \textcolor{orange}{Describe a scenario} where a GPT language model could be used for task completion. \\
      \cdashline{2-3}
      & 2.2 & \textit{(Describe the 'situation.')} \\
      & & \textcolor{orange}{Describe a situation} where you had to demonstrate teamwork. \\
      & & \textcolor{orange}{Describe a situation} where body language can help facilitate understanding. \\
      & & \textcolor{orange}{Describe a situation} where the use of solar energy is beneficial. \\
      \hline
      \multicolumn{3}{l}{\textit{layer8-4818,5557,3221,6052: Activated when the instruction requires providing something.}} \\
      & 1.6 & \textit{(The quantity is three)} \\
      & & \textcolor{orange}{Provide three} example sentences that use the word “redundant” \\
      & & \textcolor{orange}{Provide three} steps to solve a particular problem. How to create a budget \\
      & & \textcolor{orange}{Provide three} example words for the following category: fruits \\
      \cdashline{2-3}
      & 1.1 & \textit{(The quantity is four)} \\
      & & \textcolor{orange}{Provide four} examples of data visualizations. \\
      & & \textcolor{orange}{Provide four} key advantages of using a cloud-based system \\
      & & \textcolor{orange}{Provide four} ideas to boost employee morale. \\
      \hline
    \end{tabular}
    \caption{Case study of activation values.}
    \label{tab:activation_value_cases}
\end{table*}
\renewcommand{\arraystretch}{1}

\begin{table*}[t]
  \centering
  \begin{tabular}{lccccccc}
    \hline
    \textbf{Method} & \textbf{MMLU} & \textbf{GSM} & \textbf{HellaSwag} & \textbf{TruthfulQA} & \textbf{ARC-C} & \textbf{CodeX} & \textbf{Imp.} \\
    \hline
    w/o Neuron Activation & 47.17 & 17.97 & 80.62 & 56.43 & 55.18 & 16.52 & 0.93\% \\
    w/o Filtering & 47.64 & 16.60 & 80.88 & 56.06 & 54.52 & 17.26 & 0.40\% \\
    w/o Complexity-Priority & 45.12 & 17.29 & 81.16 & 59.61 & 59.87 & 16.46 & 2.02\% \\
    w/o Activation Values & 47.42 & 18.35 & 81.07 & 60.79 & 56.52 & 16.77 & 3.43\% \\
    \hdashline
    $\rm MADS_{3B}$ &	47.46 &	19.71 &	81.31 &	57.16 &	56.86 &	18.05 &	5.10\% \\
    \hline
  \end{tabular}
  \caption{Ablation study results on the four components of our framework. \textbf{w/o Neuron Activation}: replacing neuron activation states with hidden states, specifically the final output representation of each layer, for tag extraction. \textbf{w/o Filtering}: removing low-frequency activation filtering ($\theta_{base}=0$). \textbf{w/o Complexity-Priority}: randomly selecting one instruction per activation tag instead of using complexity-priority selection. \textbf{w/o Activation Values}: selecting one instruction per unique tag instead of per $(tag, v_{tag})$ pair.}
  \label{tab:ablation}
\end{table*}

\subsubsection{Ablation Study}
\label{sec:ablation}
To verify the effectiveness of the three key components in our framework, we conducted ablation studies by setting up the following four experiments:
(1) \textbf{Effect of Activation Tags Extracting}: We replaced the extraction of neuron activation states with the extraction of hidden states, specifically the final output representation of each layer, to verify the role of neuron activation states.
(2) \textbf{Effect of Activation Tags Filtering}: We did not filter low-frequency activations, specifically setting the value of $\theta_{base}$ to 0. In this case, the data volume of the core set is 81.87\%.
(3) \textbf{Effect of Full-Coverage Core Set Selection with Complexity Priority}: For each activation tag, instead of selecting the instruction containing the most different tags, we randomly selected one instruction from the instructions containing that tag.
(4) \textbf{Effect of Activation Values}: We ignore activation values during core set selection, selecting one instruction with the highest complexity for each unique $tag$ instead of for each $(tag, v_{tag})$ pair.

The experimental results are shown in Table~\ref{tab:ablation}.
From the results, we can observe that:
(1) Replacing neuron activation states with hidden states results in a significant decrease in model performance, with the average improvement dropping to 0.93\%.
This indicates that neuron activation states can better represent the features of the data compared to hidden states.
(2) When low-frequency activations are not filtered, although the core set size increases to 81.87\%, the model performance is the worst, with an average improvement of only 0.40\%.
This suggests that low-frequency activations may be noise, and retaining them introduces a large amount of redundant or low-quality data.
(3) Randomly selecting instructions for each activation tag results in an average improvement of 2.02\%, which is lower than the performance of MADS with 5.10\%.
This demonstrates that selecting instructions with higher complexity, which contain more diverse activation tags, is more effective for constructing a high-quality core set.
(4) When activation values are ignored, the average improvement decreases to 3.43\%, which is 1.67\% lower than the full MADS method.
This indicates that treating $(tag, v_{tag})$ pairs as distinct activation patterns enables finer-grained data partitioning, thereby improving the diversity and quality of the selected core set.

\subsubsection{Analysis of $v_{tag}$ Rounding Granularity}
\label{sec:vtag_analysis}
The rounding operation $\text{Round}(v_{tag}, 1)$ in Algorithm~\ref{alg:core-set} is central to the $(tag, v_{tag})$ grouping mechanism.
The choice of rounding precision directly controls the granularity of this sub-grouping.
To analyze how rounding precision affects the grouping structure and downstream task performance, we compare three settings: integer precision (0 decimal places), one decimal place (our default), and two decimal places.
All experiments use Llama-3.2-3B-Instruct for activation extraction and fine-tune Llama-2-7B with a 15\% core set from Alpaca-GPT4.

Table~\ref{tab:vtag_granularity} reports, for each precision setting, the total number of unique activation tags retained in the core set, the average number of distinct $v_{tag}$ values per unique activation tag, and the $\theta_{base}$ required to achieve a 15\% core set size.
At integer precision, each unique activation tag has on average only 1.25 distinct integer $v_{tag}$ value.
At two decimal places, the high precision creates many fine-grained sub-groups per tag, requiring an aggressive filtering threshold of $\theta_{base} = 108$ to reach the 15\% target; this over-filtering leaves only 817 activation tags, resulting in insufficient coverage of the activation space.
Compared to the extremes of integer and two-decimal precision, using one decimal place achieves a more reasonable balance: $\theta_{base} = 17$ retains 5253 diverse activation tags, with each activation tag split into an average of 3.28 sub-groups, providing fine-grained intensity discrimination while preserving broad activation-space coverage.

\begin{table*}[!t]
  \centering
  \begin{tabular}{lccc}
    \hline
    \textbf{Precision} & \textbf{\#Act.\ Patterns} & \textbf{Avg.\ $v_{tag}$ per Tag} & \textbf{$\theta_{base}$} \\
    \hline
    Integer & 23475 & 1.25 & 4 \\
    1 decimal (ours) & 5253 & 3.28 & 17 \\
    2 decimal & 817 & 16.22 & 108 \\
    \hline
  \end{tabular}
  \caption{Statistics of $(tag, v_{tag})$ patterns under different $v_{tag}$ rounding precisions, with a 15\% core set from Alpaca-GPT4. ``\#Act.\ Patterns'' denotes the total number of unique activation tags in the final core set. ``Avg.\ $v_{tag}$ per Tag'' denotes the average number of distinct $v_{tag}$ values per unique activation tag type. ``$\theta_{base}$'' denotes the uniform filtering threshold applied to each group to achieve the 15\% target core set size.}
  \label{tab:vtag_granularity}
\end{table*}

Table~\ref{tab:vtag_perf} presents the downstream task performance under each precision setting.
Integer precision achieves 1.60\% average improvement, confirming that when the grouping granularity is excessively coarse, the activation value information provides limited benefit.
Two decimal places yields only 4.76\% improvement, as the overly fine-grained grouping of $v_{tag}$ leaves too few activation tags to ensure diverse coverage.
One decimal place achieves the best performance of 5.10\%. 
This demonstrates that such level of granularity establishes an optimal balance: it captures meaningful variations in activation intensity for fine-grained sub-group diversity, while also preventing excessive pattern fragmentation, thereby preserving a comprehensive and representative coverage of the activation space.

\begin{table*}[!t]
  \centering
  \begin{tabular}{lccccccc}
    \hline
    \textbf{Precision} & \textbf{MMLU} & \textbf{GSM} & \textbf{HellaSwag} & \textbf{TruthfulQA} & \textbf{ARC-C} & \textbf{CodeX} & \textbf{Imp.} \\
    \hline
    Integer & 47.77 & 19.03 & 81.12 & 60.71 & 56.52 & 14.39 & 1.60\% \\
    1 decimal (ours) & 47.46 & 19.71 & 81.31 & 57.16 & 56.86 & 18.05 & 5.10\% \\
    2 decimal & 46.23 & 17.36 & 80.96 & 61.54 & 56.86 & 18.96 & 4.76\% \\
    \hline
  \end{tabular}
  \caption{Performance of Llama-2-7B fine-tuned with 15\% core sets selected under different $v_{tag}$ rounding precisions.}
  \label{tab:vtag_perf}
\end{table*}

\subsubsection{Analysis of Length Bias}
\label{sec:length_bias}
One potential concern is whether our greedy algorithm exhibits length bias, i.e., whether it simply selects longer instructions because they naturally activate more neurons.
To investigate this, we analyzed the instruction length distribution of the core set selected by $\rm MADS_{3B}$ and compared it with the two best-performing baselines, SelectIT and NUGGETS.

\begin{figure}[!t]
    \centering
    \includegraphics[width=0.98\columnwidth]{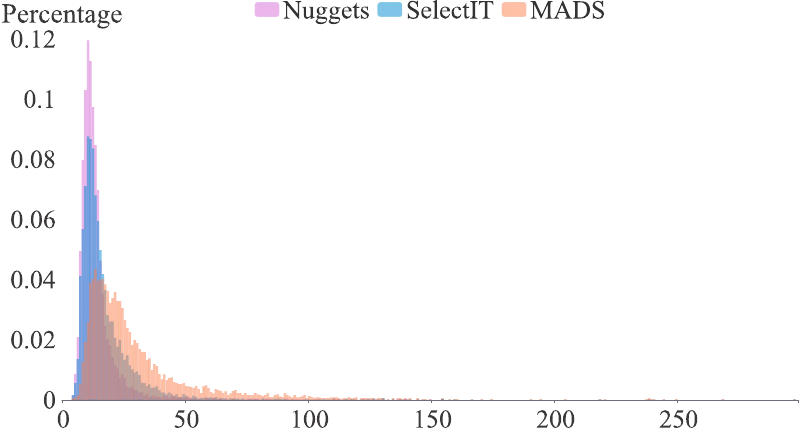}
    \caption{Frequency distribution of instruction lengths (measured in tokens) for core sets selected by three methods: $\rm MADS_{3B}$, SelectIT, and NUGGETS.
    The x-axis represents instruction length, and the y-axis represents frequency count.
    While $\rm MADS_{3B}$ includes some longer instructions, the majority of selected instructions remain within moderate length ranges similar to the baselines.
    The distributions show that MADS does not exclusively favor long instructions but maintains balanced length diversity.}
    \label{fig:instruction_length}
\end{figure}

Figure~\ref{fig:instruction_length} presents the frequency distribution of instruction lengths (measured in tokens) for the three methods.
As shown in the figure, although $\rm MADS_{3B}$ includes some longer instructions, the majority of selected instructions still fall within a moderate length range.
Table~\ref{tab:length_stats} provides detailed statistics of the instruction lengths.
The results show that while $\rm MADS_{3B}$ has a higher average and median instruction length compared to SelectIT and NUGGETS, the mode values are similar across all methods, indicating that the most frequently selected instructions are of comparable length.
Importantly, as shown in Figure~\ref{fig:instruction_length}, the length distributions of all three methods remain concentrated in similar ranges, with the majority of instructions containing fewer than 50 tokens.
The minimum instruction length for all methods is identical.
These experimental results indicate that while our method tends to select more complex instructions, it does not completely overlook shorter instructions.
Additionally, compared to the baselines, the instruction lengths we select are more balanced.

While the above descriptive statistics show that MADS does not exclusively select long instructions, a natural concern is whether the instruction complexity metric $C(ins) = |T^f(ins)|$ simply acts as a proxy for instruction length, since longer instructions contain more tokens and thus have more opportunities to activate neurons.
To disentangle the effects of length and activation complexity on MADS's selection behavior, we perform a logistic regression analysis on all instructions in the Alpaca-GPT4 dataset.
For each instruction $ins$, we compute two features: (1) the instruction length $L(ins)$ (number of tokens), and (2) the activation complexity $C(ins) = |T^f(ins)|$ (number of distinct activation tags). 
The binary outcome variable indicates whether the instruction is selected by MADS ($P(ins)=1$) or not ($P(ins)=0$). 
We fit the following logistic regression model:
\begin{equation}
    \label{eq:logistic_regression}
    P(ins) = \sigma(\beta_0 + \beta_1 \cdot L(ins) + \beta_2 \cdot C(ins))
\end{equation}
where $\sigma(\cdot)$ denotes the sigmoid function.

The regression results are presented in Table~\ref{tab:logistic_regression}.
Both instruction length ($\beta_1 = 0.042$, $p < 0.001$) and activation complexity ($\beta_2 = 0.161$, $p < 0.001$) are statistically significant predictors. 
Notably, if MADS merely selected longer instructions, the complexity coefficient would become non-significant when length is controlled; instead, it remains highly significant, confirming that complexity contributes independently to selection.
More importantly, the standardized coefficients reveal that activation complexity is a much stronger predictor than length. 
The effect size for complexity is 1.041, which is $1.9$ times larger than the 0.551 for length, demonstrating that complexity is the dominant factor driving MADS selection.
These results provide evidence that MADS captures semantic complexity beyond merely reflecting instruction length.

\begin{table*}[!t]
  \centering
  \begin{tabular}{lccccc}
    \hline
    \textbf{Method} & \textbf{Min} & \textbf{Max} & \textbf{Mean} & \textbf{Median} & \textbf{Mode} \\
    \hline
    SelectIT & 4 & 309 & 16.66 & 13 & 10 \\
    NUGGETS & 4 & 309 & 13.71 & 12 & 10 \\
    $\rm MADS_{3B}$ & 4 & 518 & 28.99 & 22 & 13 \\
    \hline
  \end{tabular}
  \caption{Statistics of instruction token counts for different core set selection methods.}
  \label{tab:length_stats}
\end{table*}

\begin{table}[!t]
  \centering
  \begin{tabular}{lcc}
    \hline
    \textbf{Variable} & \textbf{Coef.} & \textbf{Std. Coef.} \\
    \hline
    Intercept ($\beta_0$) & $-6.067^{***}$ & $-2.127$  \\
    Length ($\beta_1$) & $0.042^{***}$ & $0.551$ \\
    Complexity ($\beta_2$) & $0.161^{***}$ & $1.041$ \\
    \hline
    \multicolumn{3}{l}{\small Pseudo $R^2 = 0.243$; $N = 52{,}002$; $^{***}p < 0.001$} \\
    \hline
  \end{tabular}
  \caption{Logistic regression results predicting MADS selection. 
  ``Coef.'' denotes the original-scale coefficient, ``Std.\ Coef.'' denotes the standardized coefficient for effect size comparison.
   Pseudo $R^2$ measures the proportion of variance explained by the model, where higher values indicate better explanatory power. 
   The result of $^{***}p < 0.001$ indicates that the probability of observing such results under the null hypothesis is less than 0.1\%.
   }
  \label{tab:logistic_regression}
\end{table}

To further validate whether MADS favors higher-quality long instructions rather than arbitrary lengthy instructions under the influence of instruction complexity, we conduct a length-controlled ablation study.
We construct a length-matched random set by stratified sampling from the Alpaca-GPT4 dataset, ensuring its instruction-length distribution fully aligns with that of the $\rm MADS_{3B}$ core set.
As shown in Table~\ref{tab:length_controlled}, this length-matched random set achieves only 2.28\% average improvement, substantially lower than the 5.10\% achieved by $\rm MADS_{3B}$.
This result demonstrates that MADS selects higher-quality long instructions rather than merely long ones.

\begin{table*}[!t]
  \centering
  \begin{tabular}{lccccccc}
    \hline
    \textbf{Method} & \textbf{MMLU} & \textbf{GSM} & \textbf{HellaSwag} & \textbf{TruthfulQA} & \textbf{ARC-C} & \textbf{CodeX} & \textbf{Imp.} \\
    \hline
    Length-Matched & 46.08 & 16.91 & 81.41 & 58.26 & 58.86 & 17.32 & 2.28\% \\
    $\rm MADS_{3B}$ & 47.46 & 19.71 & 81.31 & 57.16 & 56.86 & 18.05 & 5.10\% \\
    \hline
  \end{tabular}
  \caption{Length-controlled ablation on Llama-2-7B. "Length-Matched" refers to a random set created by stratified sampling from the Alpaca-GPT4 dataset, ensuring its instruction-length distribution fully aligns with that of the $\rm MADS_{3B}$ core set.}
  \label{tab:length_controlled}
\end{table*}

\subsubsection{Computational Cost Analysis}
\label{sec:computational_cost}
In addition to downstream task performance, computational efficiency during data selection is an important practical consideration. 
In this section, we extract a 15\% core set from the Alpaca-GPT4 dataset, and the experiment for each method is independently conducted on a single NVIDIA A800 GPU.
Table~\ref{tab:computational_cost} summarizes the key computational factors across different methods.

\begin{table*}[!t]
  \centering
  \setlength{\tabcolsep}{3pt}
  \small
  \begin{tabular}{l|c|c|c|cc|cc|cc}
    \hline
    \multirow{2}{*}{\textbf{Method}} & \multirow{2}{*}{\textbf{GPT}} & \multirow{2}{*}{\textbf{Warmup}} & \multirow{2}{*}{\textbf{Pre-def.}} & \multicolumn{6}{c}{\textbf{Representation Extraction}} \\
    \cline{5-10}
    & & & & \textbf{LLM} & \textbf{\#Fwd} & \textbf{Small Model} & \textbf{\#Fwd} & \textbf{Time} & \textbf{Mem.} \\
    \hline
    DEITA & \checkmark & \checkmark & - & Llama-1-7B & 2 & - & - & 4h45m & 28GB \\
    MoDS & - & \checkmark & - & Llama-2-7B & 1 & DeBERTa (435M) & 2 & 87h25m & 70GB \\
    IFD & - & \checkmark & - & Llama-2-7B & 2 & - & - & 1h11m & 15GB \\
    NUGGETS & - & - & \checkmark & Llama-2-7B & $m$ & - & - & 223h45m & 69GB \\
    ClusterClip & - & - & \checkmark & - & - & JinaBERT (137M) & 1 & 2m20s & 5GB \\
    SelectIT & - & - & - & Llama-2-7B/13B/70B & 15 & - & - & -$^\ddagger$ & -$^\ddagger$ \\
    InsTag & \checkmark & \checkmark & - & Llama-2-7B & 1 & - & - & 29m57s & 74GB \\
    \hdashline
    MADS & - & - & - & Llama-3.2-3B & 1 & - & - & 1h09m & 15GB \\
    \hline
  \end{tabular}
  \caption{Comparison of computational requirements for data selection methods. 
  ``GPT'' indicates whether ChatGPT/GPT-4 annotation is required. 
  ``Warmup'' indicates whether model training is required before data selection. 
  ``Pre-def.'' indicates whether task categories or cluster numbers must be specified in advance. 
  ``LLM'' shows the large language model used (either directly or as a base for training) for representation extraction or scoring. 
  ``\#Fwd'' denotes the number of forward passes required per instruction, where $m$ denotes the number of predefined tasks in NUGGETS. 
  ``Small Model'' indicates additional small models required, with parameter counts in parentheses. 
  Time and memory are measured on a single NVIDIA A800 GPU using each method's official implementation, reflecting actual runtime rather than theoretical minimum (as some methods implement parallelization while others do not). 
  $^\ddagger$SelectIT requires 7B, 13B, and 70B models with 5 forward passes each, exceeding our available computational resources.}
  \label{tab:computational_cost}
\end{table*}

We briefly describe each baseline method's computational requirements:
\begin{itemize}[topsep=0pt, partopsep=0pt, itemindent=0pt]
    \setlength{\itemsep}{0pt}
    \item \textbf{DEITA}~\citep{liu2023makes} employs ChatGPT to score instruction complexity and response quality, then trains a scoring model based on Llama-1-7B. Data selection requires two forward passes: one for complexity scoring and one for quality scoring.
    \item \textbf{MoDS}~\citep{du2023mods} employs a reward model \footnote{\url{https://huggingface.co/OpenAssistant/reward-model-deberta-v3-large-v2}} for both quality evaluation and necessity evaluation. The method first uses the reward model to filter high-quality data (one DeBERTa forward pass), then fine-tunes Llama-2-7B on seed data to obtain an initial model. Finally, it uses the initial model to generate responses for all high-quality instructions (one LLM forward pass) and applies the reward model again to identify necessary data (another DeBERTa forward pass).
    \item \textbf{IFD}~\citep{li2024quantity} requires a warmup model fine-tuned on a seed dataset, then computes Instruction-Following Difficulty scores by comparing losses with and without instruction context, requiring two forward passes per instruction.
    \item \textbf{NUGGETS}~\citep{li2023one} designs a one-shot learning metric using Llama-2-7B. For each candidate instruction, it requires $m$ forward passes to compute one-shot scores on each of the $m$ predefined tasks. In our experiments, we use the official implementation with $m=100$ tasks sampled from Alpaca-GPT4 dataset via K-Means clustering.
    \item \textbf{ClusterClip}~\citep{shao-etal-2024-balanced} uses JinaBERT \footnote{\url{https://huggingface.co/jinaai/jina-embeddings-v2-base-en}} to extract text embeddings and applies k-means clustering, requiring pre-specification of the number of clusters.
    \item \textbf{SelectIT}~\citep{liu2024selectit} computes uncertainty-based metrics across three granularities (token, sentence, model) using multiple LLMs (7B, 13B, 70B), requiring 15 forward passes per instruction (5 passes per model for uncertainty estimation).
    \item \textbf{InsTag}~\citep{lu2023instag} uses ChatGPT to generate semantic tags for instructions, then trains a tagging model based on Llama-2-7B for tag prediction, requiring a single forward pass per instruction.
\end{itemize}

From Table~\ref{tab:computational_cost}, we analyze the computational efficiency of MADS. 
In terms of both time and memory consumption, only ClusterClip outperforms our method, as it relies solely on a small embedding model without requiring any LLM inference.
Regarding time efficiency alone, InsTag achieves faster processing than MADS, benefiting from its pre-trained tagging model that requires only a single forward pass for tag prediction.
However, overall, MADS demonstrates competitive computational efficiency compared to most baseline methods while offering several practical advantages:
(1) \textbf{No external API dependency}: Unlike DEITA and InsTag, MADS does not require ChatGPT annotation, eliminating API costs and potential data privacy concerns.
(2) \textbf{No warmup training}: Unlike DEITA, MoDS, and IFD, MADS operates directly on the original dataset without requiring preliminary model training, significantly reducing total computation time.
(3) \textbf{No pre-defined categories}: Unlike NUGGETS and ClusterClip, MADS does not require specifying task categories or cluster numbers in advance, as data categories are naturally derived from neuron activation patterns of LLMs.
(4) \textbf{Single forward pass}: MADS requires only one forward pass per instruction using a 3B-parameter model, which is more efficient than NUGGETS and SelectIT.

\subsubsection{Redundancy Quantification: Activation Tags vs.\ Text Embeddings}
\label{sec:redundancy}
A natural question is whether the optimization landscape induced by activation tags meaningfully differs from that of embedding-based coverage methods.
To investigate this, we conduct a redundancy quantification analysis comparing the core sets selected by MADS and ClusterClip~\citep{shao-etal-2024-balanced}, a representative embedding-based coverage method that uses JinaBERT to extract text embeddings followed by k-means clustering.
We evaluate redundancy from two complementary perspectives:
(1) \textbf{Embedding-space redundancy}: We use JinaBERT to extract text embeddings for each instruction in both core sets and compute the average pairwise cosine similarity, where lower similarity indicates less redundancy and greater diversity in the embedding space.
(2) \textbf{Activation-space coverage}: We use Llama-3.2-3B-Instruct to extract activation tags for each instruction in both core sets and count the total number of distinct activation tags covered, where a higher count indicates broader coverage of the model's internal feature space.

\begin{table}[!t]
  \centering
  \begin{tabular}{lcc}
    \hline
    \textbf{Method} & \textbf{Avg.\ Cosine Sim.} & \textbf{\#Act.\ Tags} \\
    \hline
    ClusterClip & 0.2414 & 425K \\
    $\rm MADS_{3B}$ & 0.2508 & 757K \\
    \hline
  \end{tabular}
  \caption{Redundancy quantification of core sets selected by ClusterClip and $\rm MADS_{3B}$. ``Avg.\ Cosine Sim.'' denotes the average pairwise cosine similarity of JinaBERT text embeddings within each core set (lower indicates less redundancy). ``\#Act.\ Tags'' denotes the total number of distinct activation tags covered by each core set (higher indicates broader coverage).}
  \label{tab:redundancy}
\end{table}

The results are presented in Table~\ref{tab:redundancy}.
We observe that the two core sets exhibit comparable embedding-space redundancy, with average cosine similarities of 0.2414 (ClusterClip) and 0.2508 ($\rm MADS_{3B}$).
This near-parity is expected: ClusterClip explicitly optimizes for embedding-space diversity via k-means clustering on JinaBERT embeddings, so it naturally achieves low redundancy in that space.
The fact that MADS achieves a similar level of embedding-space diversity \emph{without} directly optimizing for it suggests that activation-based selection implicitly induces comparable textual diversity.

More importantly, the two methods differ dramatically in activation-space coverage: MADS covers 757K distinct activation tags, which is \textbf{78.1\%} more than the 425K tags covered by ClusterClip.
Text embeddings from models like JinaBERT encode general semantic similarity, but they do not reflect the fine-grained feature distinctions that arise within the LLM's own representational space.
This broader activation-space coverage provides a plausible explanation for the superior downstream performance of MADS over ClusterClip (Table~\ref{tab:main}): by covering more of the LLM's internal feature space, the fine-tuned model is exposed to a more comprehensive set of training signals, leading to more balanced and effective capability improvement.

\section{Conclusion}
In this paper, we introduce MADS, which uses the neuron activation states during LLMs' inference process as the data tags, to select a diverse instruction fine-tuning dataset, achieving better fine-tuning performance with only part of training data.
Our method fully utilizes the inherent ability of LLMs to distinguish instructions with different features, and achieves the diversity and coverage of core set selection through LLMs' self-guided manner.
We evaluated MADS on multiple benchmarks, and the results indicate that MADS comprehensively enhances the performance of LLMs across various downstream tasks.

\section*{Limitations}
Despite achieving effective diverse data selection to enhance LLM performance, our method has several limitations.
The frequency filtering threshold in MADS is a hyper-parameter that requires manual setting. Future work could explore automatic threshold determination based on the distribution of activation tags, inspired by~\citet{xiao-etal-2025-finding}.
Additionally, the Full-Coverage Core Set Selection with Complexity Priority algorithm used in MADS for core set selection is a greedy algorithm.
Future work could explore other activation tags-based core set selection algorithms to enhance the performance of LLMs.
Furthermore, our experiments indicate that different layers excel in different tasks.
Future work could investigate whether the MADS method can be extended to task-specific data selection, allowing for the selection of the optimal core set for specific tasks.

\bibliography{custom}

@article{achiam2023gpt,
  author = {Achiam, J. and Adler, S. and Agarwal, S. and Ahmad, L. and Akkaya, I. and Aleman, F. L. and Almeida, D. and Altenschmidt, J. and Altman, S. and Anadkat, S. and others},
  title = {{GPT}-4 technical report},
  journal = {arXiv preprint arXiv:2303.08774},
  year = {2023}
}

@article{jiang2023mistral,
  author = {Jiang, A. Q. and Sablayrolles, A. and Mensch, A. and Bamford, C. and Chaplot, D. S. and Casas, D. de las and Bressand, F. and Lengyel, G. and Lample, G. and Saulnier, L. and others},
  title = {Mistral 7B},
  journal = {arXiv preprint arXiv:2310.06825},
  year = {2023}
}

@article{dubey2024llama,
  author = {Dubey, A. and Jauhri, A. and Pandey, A. and Kadian, A. and Al-Dahle, A. and Letman, A. and Mathur, A. and Schelten, A. and Yang, A. and Fan, A. and others},
  title = {The {Llama} 3 herd of models},
  journal = {arXiv preprint arXiv:2407.21783},
  year = {2024}
}

@article{bai2023qwen,
  author = {Bai, J. and Bai, S. and Chu, Y. and Cui, Z. and Dang, K. and Deng, X. and Fan, Y. and Ge, W. and Han, Y. and Huang, F. and others},
  title = {{Qwen} technical report},
  journal = {arXiv preprint arXiv:2309.16609},
  year = {2023}
}

@article{ouyang2022training,
  author = {Ouyang, L. and Wu, J. and Jiang, X. and Almeida, D. and Wainwright, C. and Mishkin, P. and Zhang, C. and Agarwal, S. and Slama, K. and Ray, A. and others},
  title = {Training language models to follow instructions with human feedback},
  journal = {Advances in Neural Information Processing Systems},
  volume = {35},
  pages = {27730--27744},
  year = {2022}
}

@misc{taori2023stanford,
  author = {Taori, R. and Gulrajani, I. and Zhang, T. and Dubois, Y. and Li, X. and Guestrin, C. and Liang, P. and Hashimoto, T. B.},
  title = {Stanford {Alpaca}: An Instruction-following {LLaMA} Model},
  year = {2023},
  howpublished = {\url{https://github.com/tatsu-lab/stanford_alpaca}}
}

@inproceedings{sanhmultitask,
  author = {Sanh, V. and Webson, A. and Raffel, C. and Bach, S. H. and Sutawika, L. and Alyafeai, Z. and Chaffin, A. and Stiegler, A. and Scao, T. L. and Dey, M. and others},
  title = {Multitask prompted training enables zero-shot task generalization},
  booktitle = {International Conference on Learning Representations},
  year = {2022}
}

@inproceedings{wang2023self,
  author = {Wang, Y. and Kordi, Y. and Mishra, S. and Liu, A. and Smith, N. A. and Khashabi, D. and Hajishirzi, H.},
  title = {{Self-Instruct}: Aligning language models with self-generated instructions},
  booktitle = {Proceedings of the 61st Annual Meeting of the Association for Computational Linguistics (Volume 1: Long Papers)},
  pages = {13484--13508},
  year = {2023}
}

@article{shi2024continual,
  author = {Shi, H. and Xu, Z. and Wang, H. and Qin, W. and Wang, W. and Wang, Y. and Wang, Z. and Ebrahimi, S. and Wang, H.},
  title = {Continual learning of large language models: A comprehensive survey},
  journal = {arXiv preprint arXiv:2404.16789},
  year = {2024}
}

@article{wu2024continual,
  author = {Wu, T. and Luo, L. and Li, Y.-F. and Pan, S. and Vu, T.-T. and Haffari, G.},
  title = {Continual learning for large language models: A survey},
  journal = {arXiv preprint arXiv:2402.01364},
  year = {2024}
}

@article{zhou2024lima,
  author = {Zhou, C. and Liu, P. and Xu, P. and Iyer, S. and Sun, J. and Mao, Y. and Ma, X. and Efrat, A. and Yu, P. and Yu, L. and others},
  title = {{LIMA}: Less is more for alignment},
  journal = {Advances in Neural Information Processing Systems},
  volume = {36},
  year = {2024}
}

@inproceedings{li2024quantity,
  author = {Li, M. and Zhang, Y. and Li, Z. and Chen, J. and Chen, L. and Cheng, N. and Wang, J. and Zhou, T. and Xiao, J.},
  title = {From quantity to quality: Boosting {LLM} performance with self-guided data selection for instruction tuning},
  booktitle = {Proceedings of the 2024 Conference of the North American Chapter of the Association for Computational Linguistics: Human Language Technologies (Volume 1: Long Papers)},
  pages = {7595--7628},
  year = {2024}
}

@article{du2023mods,
  author = {Du, Q. and Zong, C. and Zhang, J.},
  title = {{MoDS}: Model-oriented data selection for instruction tuning},
  journal = {arXiv preprint arXiv:2311.15653},
  year = {2023}
}

@inproceedings{lu2023instag,
  author = {Lu, K. and Yuan, H. and Yuan, Z. and Lin, R. and Lin, J. and Tan, C. and Zhou, C. and Zhou, J.},
  title = {\#{InsTag}: Instruction tagging for analyzing supervised fine-tuning of large language models},
  booktitle = {The Twelfth International Conference on Learning Representations},
  year = {2024}
}

@article{liu2024selectit,
  author = {Liu, L. and Liu, X. and Wong, D. F. and Li, D. and Wang, Z. and Hu, B. and Zhang, M.},
  title = {{SelectIT}: Selective instruction tuning for large language models via uncertainty-aware self-reflection},
  journal = {arXiv preprint arXiv:2401.03938},
  year = {2024}
}

@inproceedings{shao-etal-2024-balanced,
  author = {Shao, Y. and Li, L. and Fei, Z. and Yan, H. and Lin, D. and Qiu, X.},
  editor = {Ku, L.-W. and Martins, A. and Srikumar, V.},
  title = {Balanced data sampling for language model training with clustering},
  booktitle = {Findings of the Association for Computational Linguistics: {ACL} 2024},
  pages = {14012--14023},
  year = {2024},
  address = {Bangkok, Thailand},
  publisher = {Association for Computational Linguistics},
  doi = {10.18653/v1/2024.findings-acl.833},
  url = {https://aclanthology.org/2024.findings-acl.833}
}

@inproceedings{devlin-etal-2019-bert,
  author = {Devlin, J. and Chang, M.-W. and Lee, K. and Toutanova, K.},
  editor = {Burstein, J. and Doran, C. and Solorio, T.},
  title = {{BERT}: Pre-training of deep bidirectional transformers for language understanding},
  booktitle = {Proceedings of the 2019 Conference of the North {A}merican Chapter of the Association for Computational Linguistics: Human Language Technologies, Volume 1 (Long and Short Papers)},
  pages = {4171--4186},
  year = {2019},
  address = {Minneapolis, Minnesota},
  publisher = {Association for Computational Linguistics},
  doi = {10.18653/v1/N19-1423},
  url = {https://aclanthology.org/N19-1423}
}

@article{chen2023maybe,
  author = {Chen, H. and Zhang, Y. and Zhang, Q. and Yang, H. and Hu, X. and Ma, X. and Yanggong, Y. and Zhao, J.},
  title = {Maybe only 0.5\% data is needed: A preliminary exploration of low training data instruction tuning},
  journal = {arXiv preprint arXiv:2305.09246},
  year = {2023}
}

@inproceedings{das2024deft,
  author = {Das, D. and Khetan, V.},
  title = {{DEFT-UCS}: Data efficient fine-tuning for pre-trained language models via unsupervised core-set selection for text-editing},
  booktitle = {Proceedings of the 2024 Conference on Empirical Methods in Natural Language Processing},
  pages = {20296--20312},
  year = {2024}
}

@article{song2024iterselecttune,
  author = {Song, J. and Liu, S. and Zhu, B. and Rao, Y.},
  title = {{IterSelectTune}: An iterative training framework for efficient instruction-tuning data selection},
  journal = {arXiv preprint arXiv:2410.13464},
  year = {2024}
}

@inproceedings{cao2023instruction,
  author = {Cao, Y. and Kang, Y. and Wang, C. and Sun, L.},
  title = {Instruction mining: Instruction data selection for tuning large language models},
  booktitle = {First Conference on Language Modeling},
  year = {2024}
}

@article{xu2023rethinking,
  author = {Xu, Y. and Yao, Y. and Huang, Y. and Qi, M. and Wang, M. and Gu, B. and Sundaresan, N.},
  title = {Rethinking the instruction quality: Lift is what you need},
  journal = {arXiv preprint arXiv:2310.07317},
  year = {2023}
}

@article{pang2024improving,
  author = {Pang, J. and Wei, J. and Shah, A. P. and Zhu, Z. and Wang, Y. and Qian, C. and Liu, Y. and Bao, Y. and Wei, W.},
  title = {Improving data efficiency via curating {LLM}-driven rating systems},
  journal = {arXiv preprint arXiv:2410.10877},
  year = {2024}
}

@inproceedings{liu2023makes,
  author = {Liu, W. and Zeng, W. and He, K. and Jiang, Y. and He, J.},
  title = {What makes good data for alignment? A comprehensive study of automatic data selection in instruction tuning},
  booktitle = {The Twelfth International Conference on Learning Representations},
  year = {2024}
}

@article{qin2024unleashing,
  author = {Qin, Y. and Yang, Y. and Guo, P. and Li, G. and Shao, H. and Shi, Y. and Xu, Z. and Gu, Y. and Li, K. and Sun, X.},
  title = {Unleashing the power of data tsunami: A comprehensive survey on data assessment and selection for instruction tuning of language models},
  journal = {Transactions on Machine Learning Research},
  year = {2024}
}

@inproceedings{joaquin2024in2core,
  author = {San Joaquin, A. and Wang, B. and Liu, Z. and Muller, P. and Asher, N. and Lim, B. Y. and Chen, N. F.},
  title = {{In2Core}: Leveraging influence functions for coreset selection in instruction finetuning of large language models},
  booktitle = {Findings of the Association for Computational Linguistics: {EMNLP} 2024},
  pages = {10324--10335},
  year = {2024},
  publisher = {Association for Computational Linguistics}
}

@article{yang2024smalltolarge,
  author = {Yang, Y. and Mishra, S. and Chiang, J. and Mirzasoleiman, B.},
  title = {{SmalltoLarge} ({S2L}): Scalable data selection for fine-tuning large language models by summarizing training trajectories of small models},
  journal = {Advances in Neural Information Processing Systems},
  volume = {37},
  pages = {83465--83496},
  year = {2024}
}

@inproceedings{pan2024g,
  author = {Pan, X. and Huang, L. and Kang, L. and Liu, Z. and Lu, Y. and Cheng, S.},
  editor = {Ku, L.-W. and Martins, A. and Srikumar, V.},
  title = {{G}-{DIG}: Towards gradient-based {DI}verse and hi{G}h-quality instruction data selection for machine translation},
  booktitle = {Proceedings of the 62nd Annual Meeting of the Association for Computational Linguistics (Volume 1: Long Papers)},
  pages = {15395--15406},
  year = {2024},
  address = {Bangkok, Thailand},
  publisher = {Association for Computational Linguistics},
  doi = {10.18653/v1/2024.acl-long.821},
  url = {https://aclanthology.org/2024.acl-long.821}
}

@inproceedings{har2005smaller,
  author = {Har-Peled, S. and Kushal, A.},
  title = {Smaller coresets for k-median and k-means clustering},
  booktitle = {Proceedings of the twenty-first annual symposium on Computational geometry},
  pages = {126--134},
  year = {2005}
}

@article{paul2021deep,
  author = {Paul, M. and Ganguli, S. and Dziugaite, G. K.},
  title = {Deep learning on a data diet: Finding important examples early in training},
  journal = {Advances in Neural Information Processing Systems},
  volume = {34},
  pages = {20596--20607},
  year = {2021}
}

@inproceedings{mirzasoleiman2020coresets,
  author = {Mirzasoleiman, B. and Bilmes, J. and Leskovec, J.},
  title = {Coresets for data-efficient training of machine learning models},
  booktitle = {International Conference on Machine Learning},
  pages = {6950--6960},
  year = {2020},
  publisher = {PMLR}
}

@article{munteanu2018coresets,
  author = {Munteanu, A. and Sohler, C. and Schwiegelshohn, C. and Woodruff, D. P. and others},
  title = {On coresets for logistic regression},
  journal = {Advances in Neural Information Processing Systems},
  pages = {6561--6570},
  year = {2018}
}

@misc{bricken2023monosemanticity,
  author = {Bricken, T. and Templeton, A. and Batson, J. and Chen, B. and Jermyn, A. and Conerly, T. and Turner, N. and Anil, C. and Denison, C. and Askell, A. and Lasenby, R. and Wu, Y. and Kravec, S. and Schiefer, N. and Maxwell, T. and Joseph, N. and Hatfield-Dodds, Z. and Tamkin, A. and Nguyen, K. and McLean, B. and Burke, J. E. and Hume, T. and Carter, S. and Henighan, T. and Olah, C.},
  title = {Towards monosemanticity: Decomposing language models with dictionary learning},
  year = {2023},
  howpublished = {\url{https://transformer-circuits.pub/2023/monosemantic-features/index.html}}
}

@misc{elhage2022superposition,
  author = {Elhage, N. and Hume, T. and Olsson, C. and Schiefer, N. and Henighan, T. and Kravec, S. and Hatfield-Dodds, Z. and Lasenby, R. and Drain, D. and Chen, C. and Grosse, R. and McCandlish, S. and Kaplan, J. and Amodei, D. and Wattenberg, M. and Olah, C.},
  title = {Toy models of superposition},
  year = {2022},
  howpublished = {\url{https://transformer-circuits.pub/2022/toy_model/index.html}}
}

@inproceedings{belinkov2018evaluating,
  author = {Belinkov, Y. and M{\`a}rquez, L. and Sajjad, H. and Durrani, N. and Dalvi, F. and Glass, J.},
  title = {Evaluating layers of representation in neural machine translation on part-of-speech and semantic tagging tasks},
  booktitle = {Proceedings of the Eighth International Joint Conference on Natural Language Processing (Volume 1: Long Papers)},
  pages = {1--10},
  year = {2017}
}

@inproceedings{peters2018dissecting,
  author = {Peters, M. E. and Neumann, M. and Zettlemoyer, L. and Yih, W.},
  title = {Dissecting contextual word embeddings: Architecture and representation},
  booktitle = {Proceedings of the 2018 Conference on Empirical Methods in Natural Language Processing},
  pages = {1499--1509},
  year = {2018}
}

@inproceedings{blevins2018deep,
  author = {Blevins, T. and Levy, O. and Zettlemoyer, L.},
  title = {Deep {RNN}s encode soft hierarchical syntax},
  booktitle = {Proceedings of the 56th Annual Meeting of the Association for Computational Linguistics (Volume 2: Short Papers)},
  pages = {14--19},
  year = {2018}
}

@article{hendrycks2020measuring,
  author = {Hendrycks, D. and Burns, C. and Basart, S. and Zou, A. and Mazeika, M. and Song, D. and Steinhardt, J.},
  title = {Measuring massive multitask language understanding},
  journal = {arXiv preprint arXiv:2009.03300},
  year = {2020}
}

@article{cobbe2021training,
  author = {Cobbe, K. and Kosaraju, V. and Bavarian, M. and Chen, M. and Jun, H. and Kaiser, L. and Plappert, M. and Tworek, J. and Hilton, J. and Nakano, R. and others},
  title = {Training verifiers to solve math word problems},
  journal = {arXiv preprint arXiv:2110.14168},
  year = {2021}
}

@article{chen2021evaluating,
  author = {Chen, M. and Tworek, J. and Jun, H. and Yuan, Q. and Ponde de Oliveira Pinto, H. and Kaplan, J. and Edwards, H. and Burda, Y. and Joseph, N. and Brockman, G. and others},
  title = {Evaluating large language models trained on code},
  journal = {arXiv preprint arXiv:2107.03374},
  year = {2021}
}

@inproceedings{zellers2019hellaswag,
  author = {Zellers, R. and Holtzman, A. and Bisk, Y. and Farhadi, A. and Choi, Y.},
  title = {{HellaSwag}: Can a machine really finish your sentence?},
  booktitle = {Proceedings of the 57th Annual Meeting of the Association for Computational Linguistics},
  pages = {4791--4800},
  year = {2019}
}

@inproceedings{lin2021truthfulqa,
  author = {Lin, S. and Hilton, J. and Evans, O.},
  title = {{TruthfulQA}: Measuring how models mimic human falsehoods},
  booktitle = {Proceedings of the 60th Annual Meeting of the Association for Computational Linguistics (Volume 1: Long Papers)},
  pages = {3214--3252},
  year = {2022}
}

@article{clark2018think,
  author = {Clark, P. and Cowhey, I. and Etzioni, O. and Khot, T. and Sabharwal, A. and Schoenick, C. and Tafjord, O.},
  title = {Think you have solved question answering? try {ARC}, the {AI2} reasoning challenge},
  journal = {arXiv preprint arXiv:1803.05457},
  year = {2018}
}

@inproceedings{li2023one,
  author = {Li, Y. and Hui, B. and Xia, X. and Yang, J. and Yang, M. and Zhang, L. and Si, S. and Chen, L.-H. and Liu, J. and Liu, T. and Huang, F. and Li, Y.},
  editor = {Ku, L.-W. and Martins, A. and Srikumar, V.},
  title = {One-shot learning as instruction data prospector for large language models},
  booktitle = {Proceedings of the 62nd Annual Meeting of the Association for Computational Linguistics (Volume 1: Long Papers)},
  pages = {4586--4601},
  year = {2024},
  address = {Bangkok, Thailand},
  publisher = {Association for Computational Linguistics},
  doi = {10.18653/v1/2024.acl-long.252},
  url = {https://aclanthology.org/2024.acl-long.252}
}

@misc{bills2023language,
  author = {Bills, S. and Cammarata, N. and Mossing, D. and Tillman, H. and Gao, L. and Goh, G. and Sutskever, I. and Leike, J. and Wu, J. and Saunders, W.},
  title = {Language models can explain neurons in language models},
  year = {2023},
  howpublished = {\url{https://openaipublic.blob.core.windows.net/neuron-explainer/paper/index.html}}
}

@article{touvron2023llama,
  author = {Touvron, H. and Martin, L. and Stone, K. and Albert, P. and Almahairi, A. and Babaei, Y. and Bashlykov, N. and Batra, S. and Bhargava, P. and Bhosale, S. and others},
  title = {{LLaMA} 2: Open foundation and fine-tuned chat models},
  journal = {arXiv preprint arXiv:2307.09288},
  year = {2023}
}

@article{peng2023instruction,
  author = {Peng, B. and Li, C. and He, P. and Galley, M. and Gao, J.},
  title = {Instruction tuning with {GPT}-4},
  journal = {arXiv preprint arXiv:2304.03277},
  year = {2023}
}

@article{zhang2025best,
  author = {Zhang, D. and Dai, Q. and Peng, H.},
  title = {The best instruction-tuning data are those that fit},
  journal = {arXiv preprint arXiv:2502.04194},
  year = {2025}
}

@article{dai2025improving,
  author = {Dai, Q. and Zhang, D. and Ma, J. W. and Peng, H.},
  title = {Improving influence-based instruction tuning data selection for balanced learning of diverse capabilities},
  journal = {arXiv preprint arXiv:2501.12147},
  year = {2025}
}

@article{zhao2025beyond,
  author = {Zhao, Y. and Du, L. and Ding, X. and Ouyang, Y. and Wang, H. and Xiong, K. and Gao, J. and Sun, Z. and Xu, D. and Qing, Y. and others},
  title = {Beyond similarity: A gradient-based graph method for instruction tuning data selection},
  journal = {arXiv preprint arXiv:2502.11062},
  year = {2025}
}

@inproceedings{zhou2025davir,
  author = {Zhou, H. and Liu, T. and Ma, Q. and Zhang, Y. and Yuan, J. and Liu, P. and You, Y. and Yang, H.},
  title = {{DAVIR}: Data selection via implicit reward for large language models},
  booktitle = {Proceedings of the 63rd Annual Meeting of the Association for Computational Linguistics (Volume 1: Long Papers)},
  pages = {9220--9237},
  year = {2025}
}

@article{hu2025donod,
  author = {Hu, J. and Yang, S. and Zhou, D. and Wu, L.},
  title = {{DONOD}: Robust and generalizable instruction fine-tuning for {LLMs} via model-intrinsic dataset pruning},
  journal = {arXiv preprint arXiv:2504.14810},
  year = {2025}
}

@inproceedings{xu2024wizardlm,
  author={Xu, Can and Sun, Qingfeng and Zheng, Kai and Geng, Xiubo and Zhao, Pu and Feng, Jiazhan and Tao, Chongyang and Lin, Qingwei and Jiang, Daxin},
  title={WizardLM: Empowering large pre-trained language models to follow complex instructions},
  booktitle={The Twelfth International Conference on Learning Representations},
  year={2024}
}

@article{xia2024rethinking,
  title={Rethinking data selection at scale: Random selection is almost all you need},
  author={Xia, Tingyu and Yu, Bowen and Dang, Kai and Yang, An and Wu, Yuan and Tian, Yuan and Chang, Yi and Lin, Junyang},
  journal={arXiv preprint arXiv:2410.09335},
  year={2024}
}

@article{luo2025inversescope,
  title={InverseScope: Scalable Activation Inversion for Interpreting Large Language Models},
  author={Luo, Yifan and Zhou, Zhennan and Dong, Bin},
  journal={arXiv preprint arXiv:2506.07406},
  year={2025}
}

@article{helff2025activationreasoning,
  title={ActivationReasoning: Logical Reasoning in Latent Activation Spaces},
  author={Helff, Lukas and H{\"a}rle, Ruben and Stammer, Wolfgang and Friedrich, Felix and Brack, Manuel and W{\"u}st, Antonia and Shindo, Hikaru and Schramowski, Patrick and Kersting, Kristian},
  journal={arXiv preprint arXiv:2510.18184},
  year={2025}
}

@article{shafran2025decomposing,
  title={Decomposing MLP Activations into Interpretable Features via Semi-Nonnegative Matrix Factorization},
  author={Shafran, Or and Geiger, Atticus and Geva, Mor},
  journal={arXiv preprint arXiv:2506.10920},
  year={2025}
}

@techreport{settles2009active,
  author = {Settles, B.},
  title = {Active learning literature survey},
  institution = {University of Wisconsin--Madison},
  number = {1648},
  year = {2009}
}

@inproceedings{sener2018active,
  title={Active Learning for Convolutional Neural Networks: A Core-Set Approach},
  author={Sener, Ozan and Savarese, Silvio},
  booktitle={International Conference on Learning Representations},
  year={2018}
}

@inproceedings{ash2019deep,
  title={Deep batch active learning by diverse, uncertain gradient lower bounds},
  author={Ash, Jordan T and Zhang, Chicheng and Krishnamurthy, Akshay and Langford, John and Agarwal, Alekh},
  booktitle={International Conference on Learning Representations},
  year={2019}
}

@inproceedings{margatina2021active,
  title={Active Learning by Acquiring Contrastive Examples},
  author={Margatina, Katerina and Vernikos, Giorgos and Barrault, Lo{\"\i}c and Aletras, Nikolaos},
  booktitle={Proceedings of the 2021 Conference on Empirical Methods in Natural Language Processing},
  pages={650--663},
  year={2021}
}

@inproceedings{xia2024less,
  title={LESS: selecting influential data for targeted instruction tuning},
  author={Xia, Mengzhou and Malladi, Sadhika and Gururangan, Suchin and Arora, Sanjeev and Chen, Danqi},
  booktitle={Proceedings of the 41st International Conference on Machine Learning},
  pages={54104--54132},
  year={2024}
}

@inproceedings{xiao-etal-2025-finding,
    title = "Finding the Sweet Spot: Preference Data Construction for Scaling Preference Optimization",
    author = "Xiao, Yao  and
      Ye, Hai  and
      Chen, Linyao  and
      Ng, Hwee Tou  and
      Bing, Lidong  and
      Li, Xiaoli  and
      Lee, Roy Ka-Wei",
    editor = "Che, Wanxiang  and
      Nabende, Joyce  and
      Shutova, Ekaterina  and
      Pilehvar, Mohammad Taher",
    booktitle = "Proceedings of the 63rd Annual Meeting of the Association for Computational Linguistics (Volume 1: Long Papers)",
    month = jul,
    year = "2025",
    address = "Vienna, Austria",
    publisher = "Association for Computational Linguistics",
    url = "https://aclanthology.org/2025.acl-long.615/",
    doi = "10.18653/v1/2025.acl-long.615",
    pages = "12538--12552",
    ISBN = "979-8-89176-251-0"
}

@article{li2025echoesbert,
  author = {Li, Michael and Subramani, Nishant},
  title = {Echoes of BERT: Do Modern Language Models Rediscover the Classical NLP Pipeline?},
  journal = {arXiv preprint arXiv:2506.02132},
  year = {2025},
  doi = {10.48550/arXiv.2506.02132},
  url = {https://arxiv.org/abs/2506.02132}
}

@inproceedings{he2024decodingprobing,
  title={Decoding probing: Revealing internal linguistic structures in neural language models using minimal pairs},
  author={He, Linyang and Chen, Peili and Nie, Ercong and Li, Yuanning and Brennan, Jonathan R},
  booktitle={Proceedings of the 2024 Joint International Conference on Computational Linguistics, Language Resources and Evaluation (LREC-COLING 2024)},
  pages={4488--4497},
  year={2024}
}

@article{jiang2024layerwisesimilarity,
  author = {Jiang, Jiachen and Zhou, Jinxin and Zhu, Zhihui},
  title = {Tracing Representation Progression: Analyzing and Enhancing Layer-Wise Similarity},
  journal = {arXiv preprint arXiv:2406.14479},
  year = {2024},
  doi = {10.48550/arXiv.2406.14479},
  url = {https://arxiv.org/abs/2406.14479}
}

@article{cunningham2023saeinterpretable,
  author = {Cunningham, Hoagy and Ewart, Aidan and Riggs, Logan and Huben, Robert and Sharkey, Lee},
  title = {Sparse Autoencoders Find Highly Interpretable Features in Language Models},
  journal = {arXiv preprint arXiv:2309.08600},
  year = {2023},
  doi = {10.48550/arXiv.2309.08600},
  url = {https://arxiv.org/abs/2309.08600}
}

@inproceedings{aligncot2023,
    title = "Aligning Large and Small Language Models via Chain-of-Thought Reasoning",
    author = "Ranaldi, Leonardo  and
      Freitas, Andre",
    editor = "Graham, Yvette  and
      Purver, Matthew",
    booktitle = "Proceedings of the 18th Conference of the European Chapter of the Association for Computational Linguistics (Volume 1: Long Papers)",
    month = mar,
    year = "2024",
    address = "St. Julian{'}s, Malta",
    publisher = "Association for Computational Linguistics",
    url = "https://aclanthology.org/2024.eacl-long.109/",
    doi = "10.18653/v1/2024.eacl-long.109",
    pages = "1812--1827"
}

@inproceedings{selfrefineit2024,
    title = "Self-Refine Instruction-Tuning for Aligning Reasoning in Language Models",
    author = "Ranaldi, Leonardo  and
      Freitas, Andre",
    editor = "Al-Onaizan, Yaser  and
      Bansal, Mohit  and
      Chen, Yun-Nung",
    booktitle = "Proceedings of the 2024 Conference on Empirical Methods in Natural Language Processing",
    month = nov,
    year = "2024",
    address = "Miami, Florida, USA",
    publisher = "Association for Computational Linguistics",
    url = "https://aclanthology.org/2024.emnlp-main.139/",
    doi = "10.18653/v1/2024.emnlp-main.139",
    pages = "2325--2347"
}

\appendix
\section{Activation Level of Neurons}
\label{appendix:appendix}
We extracted experimental results from previous work~\citep{bricken2023monosemanticity} concerning the relationship between activation levels and their corresponding text, including examples of strongly activated neurons and cases where activation values were around 1, which is the threshold we set to filter activation values, as shown in Table~\ref{table:act_val}.

\begin{table*}
  \centering
  \renewcommand{\arraystretch}{1.2}
  \begin{tabular}{>{\raggedright\arraybackslash}p{6cm}l}
    \hline
    \textbf{neurons} & \textbf{text \& activation level} \\
    \hline
    \multirow{3}{=}{fires on words related to physics and engineering} 
      & perpendicular 235nm laser \underline{beam}\textcolor{orange}{($v_{act} = 7.90$)}… \\
      & ${pm}$ are coupling \underline{constants}\textcolor{orange}{($v_{act} = 1.49$)}. The equations of… \\
      & tell these attempts were \underline{successful}\textcolor{orange}{($v_{act} = 0.69$)} in… \\
    \hdashline
    \multirow{3}{=}{High activations on programming languages, version numbers and technical specifications} 
      & be anything, and \underline{Python}\textcolor{orange}{($v_{act} = 6.2$)} has no idea… \\
      & I'll use my \underline{1TB}\textcolor{orange}{($v_{act} = 1.53$)} NAS drive… \\
      & \underline{layer}\textcolor{orange}{($v_{act} = 0.51$)} \{ name: … \\
    \hdashline
    \multirow{3}{=}{The neuron attends to variable names in source code.} 
      & \underline{var}\textcolor{orange}{($v_{act} = 5.69$)} z = y;… \\
      & \underline{out}\textcolor{orange}{($v_{act} = 1.32$)}.println();… \\
      & the particulars next \underline{time}\textcolor{orange}{($v_{act} = 0.51$)}… \\
    \hdashline
    \multirow{3}{=}{The neuron fires on math symbols, words, and notation.} 
      & $(x) = \underline{\alpha}$\textcolor{orange}{($v_{act} = 5.50$)}… \\
      & 4*-2). \underline{Is}\textcolor{orange}{($v_{act} = 1.13$)} 35 a factor of… \\
      & What is the \underline{Best}\textcolor{orange}{($v_{act} = 0.75$)} Type of TV for… \\
    \hdashline
    \multirow{3}{=}{The neuron fires on words related to analysis, measurement, and quantification.}
      & arterial bypasses \underline{depending}\textcolor{orange}{($v_{act} = 3.5$)} on the clinical stage… \\
      & players and administrators more \underline{than}\textcolor{orange}{($v_{act} = 1.19$)} AFL and tennis… \\
      & trying to get it \underline{so}\textcolor{orange}{($v_{act} = 0.95$)} that when… \\
    \hline
  \end{tabular}
  \caption{Examples of activation levels and their corresponding text}
  \label{table:act_val}
\end{table*}


\section{Additional Visualization Results}
\label{appendix:visualization}
In this section, we provide the PCA visualization and similarity heatmap for layers 20 and 27, which are not included in the main text due to space limitations.
\begin{figure*}[h]
    \centering
    \includegraphics[width=0.4\textwidth]{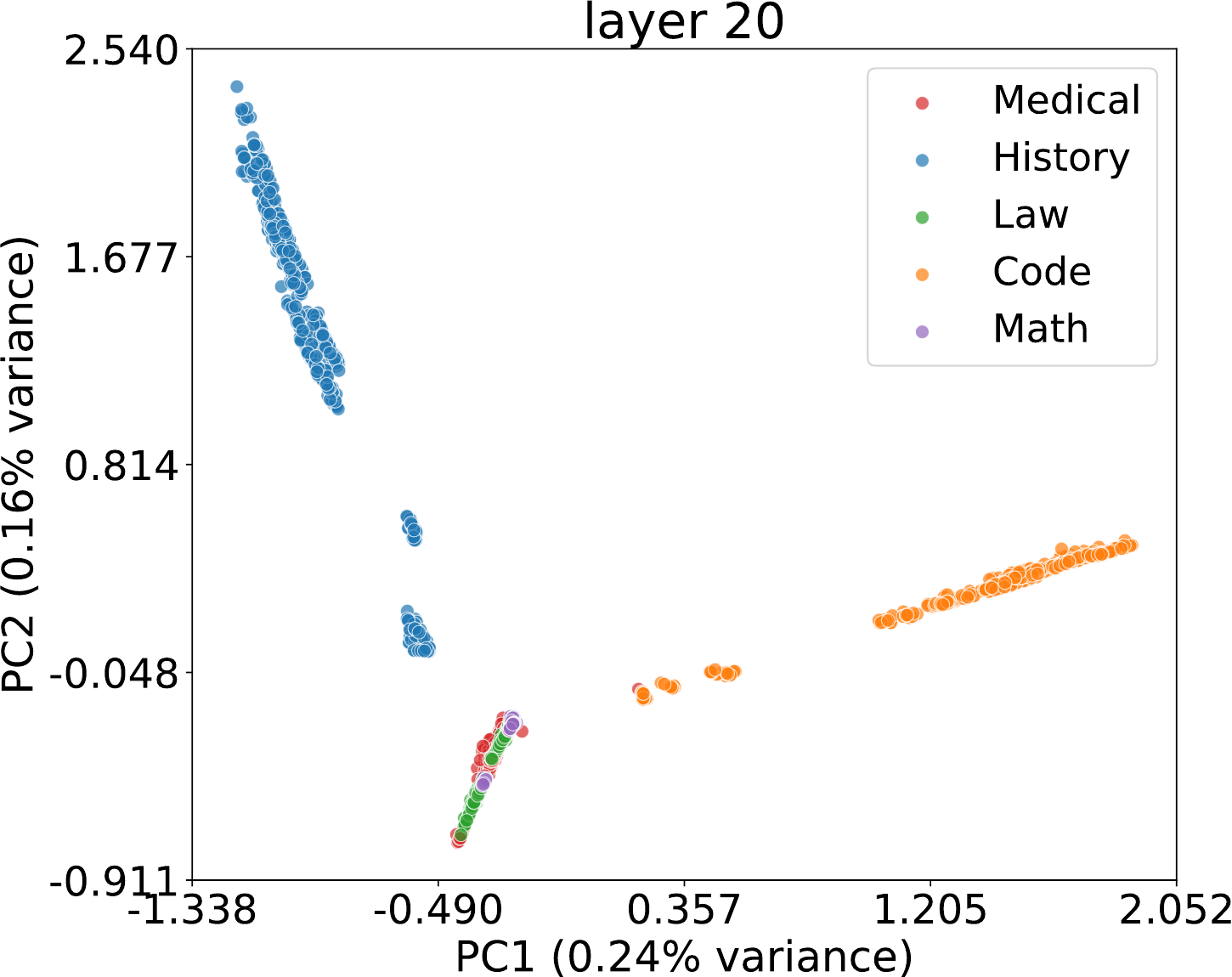}
    \includegraphics[width=0.4\textwidth]{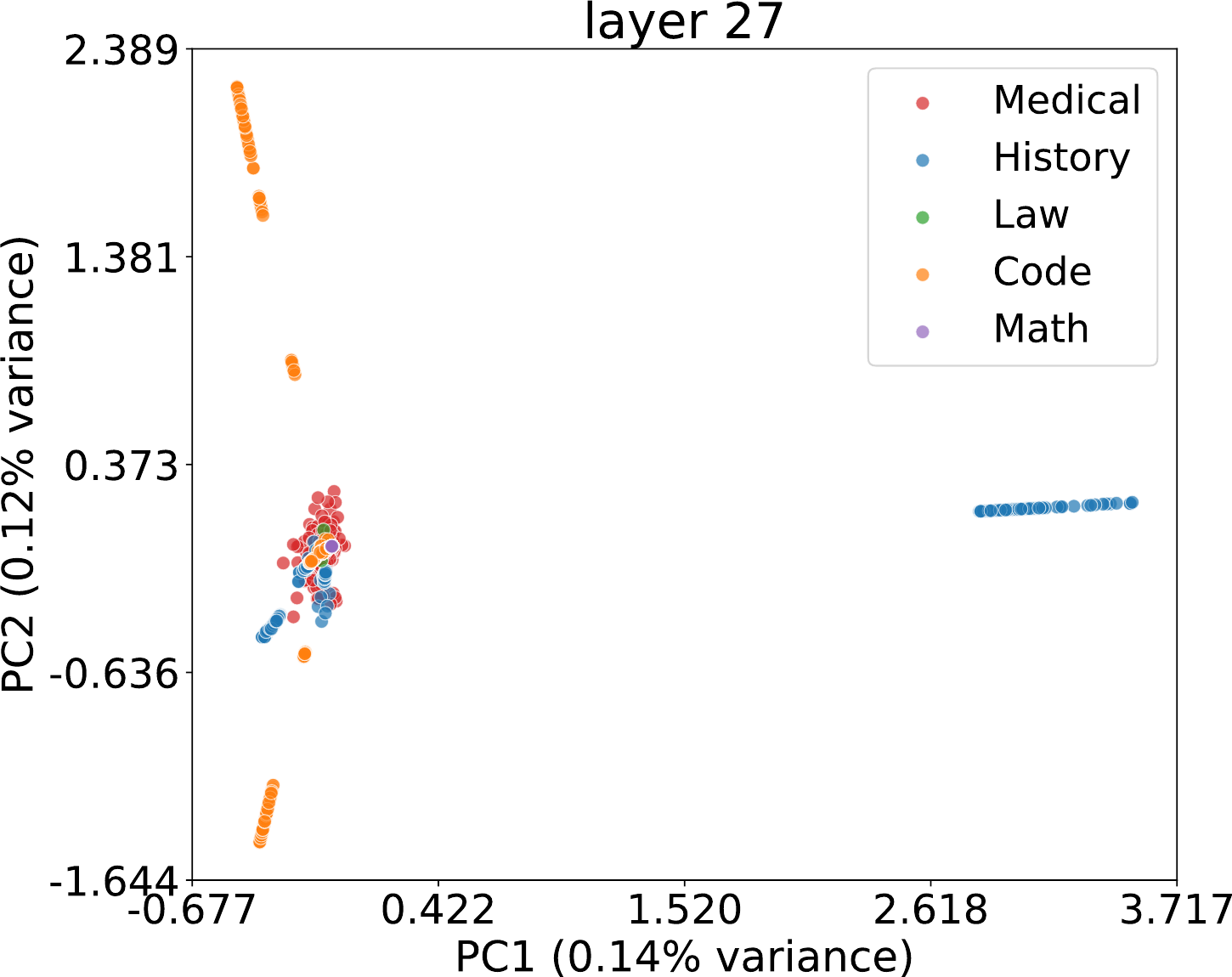}
    \caption{PCA visualization of activation tag vectors at layer 20 and 27 of Llama-3.2-3B-Instruct for five instruction categories: code (orange), math (purple), law (green), medical (red), and history (blue). 
The code and history categories demonstrate more pronounced separation from the remaining three categories.
This observation is intuitively reasonable: code-related instructions involve distinct programming syntax, while history-related queries primarily require factual recall, in contrast to the other three domains which emphasize logical reasoning processes.}
    \label{fig:pca_visualization_appendix}
\end{figure*}

\begin{figure*}[h]
    \centering
    \includegraphics[width=0.4\textwidth]{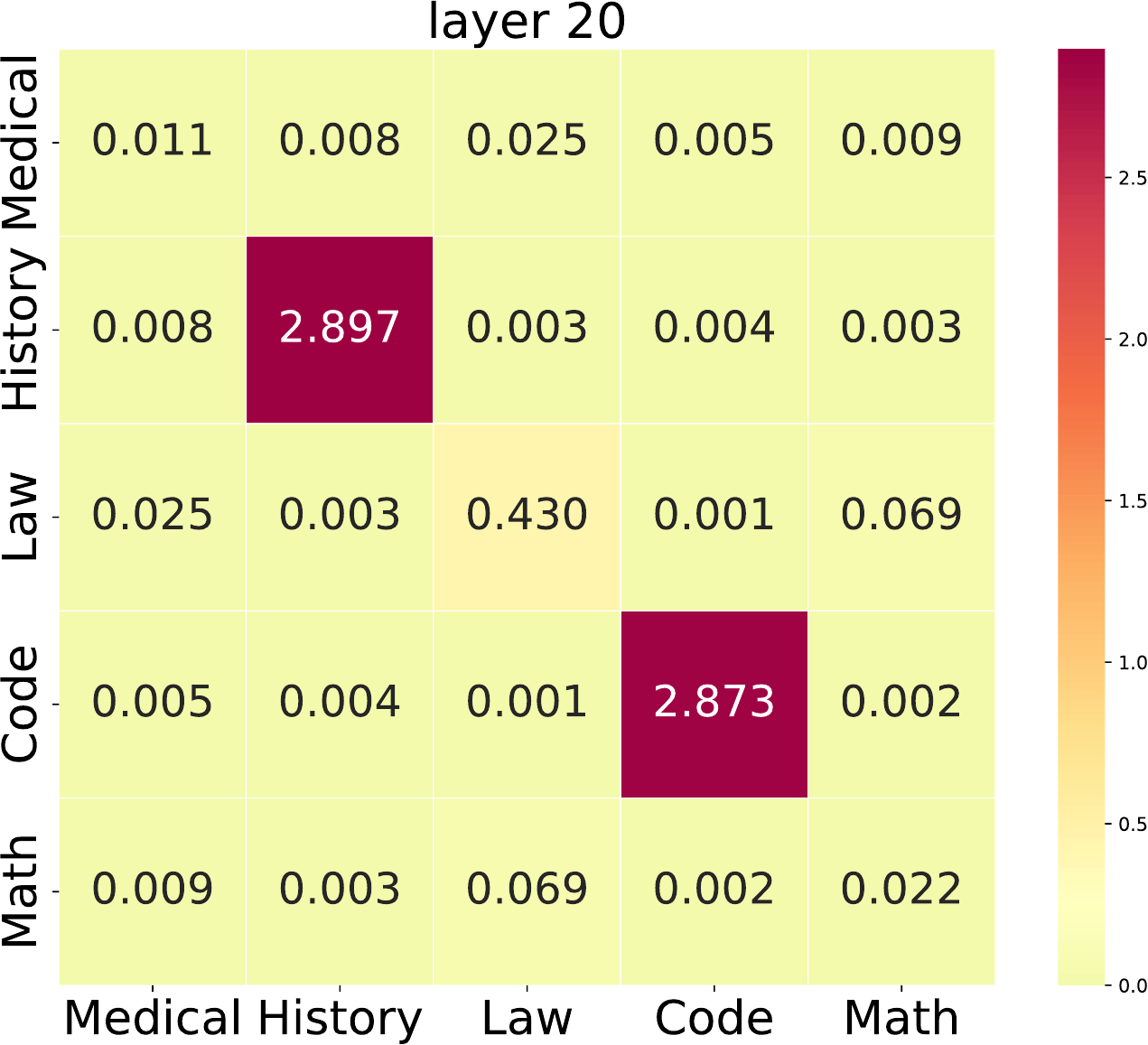}
    \includegraphics[width=0.4\textwidth]{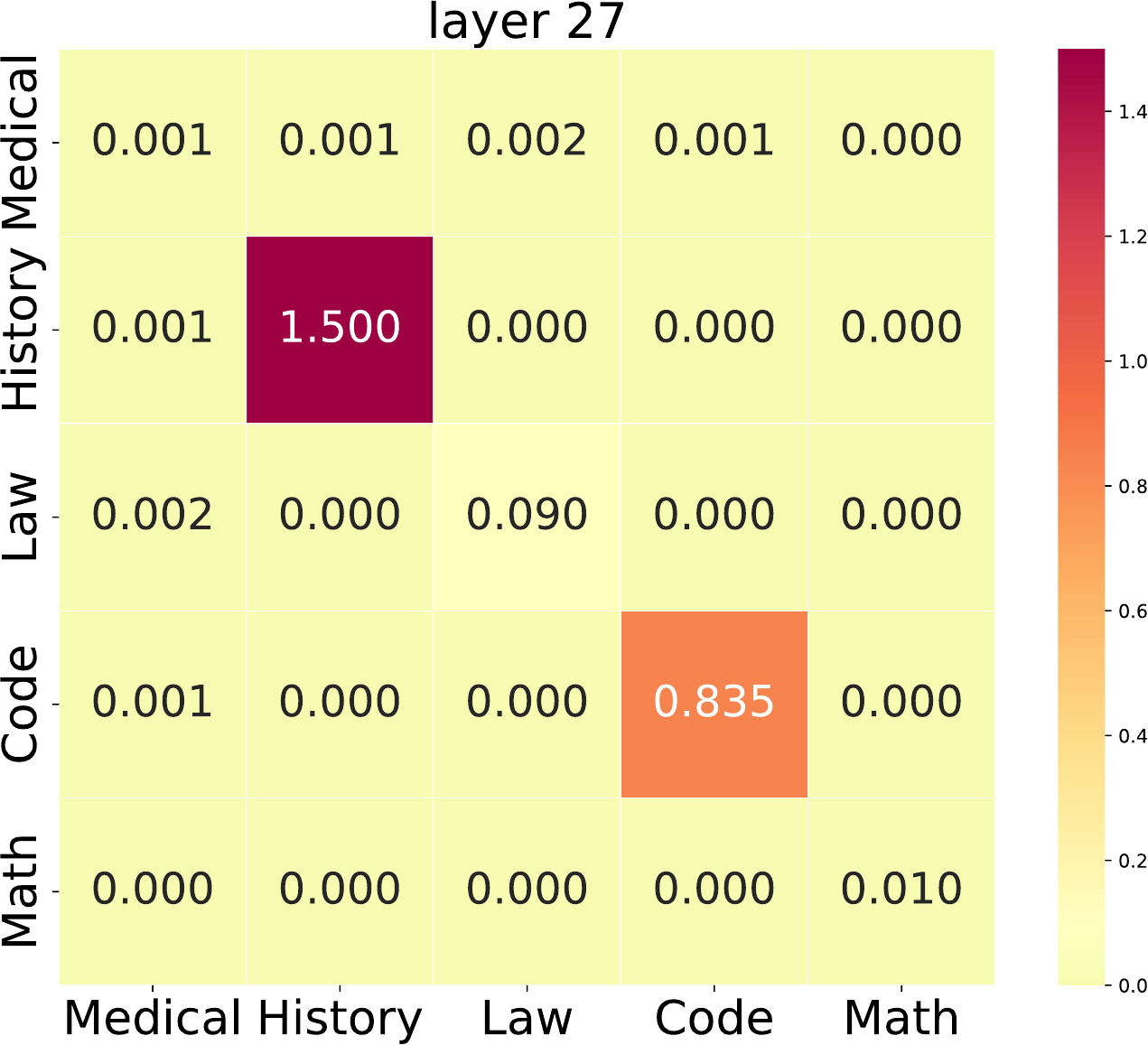}
    \caption{Heatmap of average shared activation tags between instruction categories at layers 20 and 27 of Llama-3.2-3B-Instruct.
    Consistent with Figure~\ref{fig:similarity_heatmap}, diagonal values (intra-domain) exceed off-diagonal values (cross-domain), confirming the robustness of the correlation between activation patterns and data features across deeper layers.}
    \label{fig:similarity_heatmap_appendix}
\end{figure*}

\section{Dataset-Internal Characteristics: Representative Examples}
\label{appendix:domain_examples}
Table~\ref{tab:domain_examples} presents representative instruction examples from each of the five domains used in the preliminary experiment, illustrating the dataset-internal structural differences discussed in Section~\ref{sec:preliminary_validation}.

\begin{table*}[h]
  \centering
  \renewcommand{\arraystretch}{1.3}
  \begin{tabular}{>{\raggedright\arraybackslash}p{1.5cm}
                  >{\raggedright\arraybackslash}p{3.0cm}
                  >{\raggedright\arraybackslash}p{9.5cm}}
    \hline
    \textbf{Domain} & \textbf{Internal Structure} & \textbf{Representative Instruction} \\
    \hline
    Code & Programming syntax: code blocks, \texttt{def}/\texttt{class}/\texttt{import} tokens &
    \textit{You are given a Python class method that processes a string `sval` and extracts certain keywords based on the positions stored in the `cut` list. Your task is to implement a function that takes the input string `sval` and the `cut` list, and returns the extracted keywords as a dictionary...Function signature: `def extract_keywords(sval: str, cut: List[int]) -> Dict[str, Union[int, str]] ...} \\
    \hdashline
    History & Uniform passage-based MCQ: read excerpt $\rightarrow$ select answer &
    \textit{This question refers to the following information. [Primary source excerpt on a historical event$\ldots$] Which of the following best describes the author's view? [Multiple options]} \\
    \hdashline
    Math & Arithmetic reasoning &
    \textit{If Clover walks 1.5 miles in the morning and 1.5 miles in the evening every day, how many miles does he walk in 30 days?} \newline
    \textit{(Solution: $1.5{+}1.5{=}3$ miles/day; $3{\times}30{=}90$ miles)} \\
    \hdashline
    Law & \multirow{2}{=}{Mixed task types: rule-retrieval from statutes vs.\ multi-step legal scenario reasoning} &
    \textit{[Rule-retrieval]} Congress enacts a \$100 tax on handgun sales to private individuals. Will this survive a constitutional challenge? (A) Yes, if Congress could have banned possession of handguns outright. (B) Yes, if$\ldots$ \\
    \cdashline{3-3}
    & &
    \textit{[Scenario reasoning]} As part of his defense to a murder charge, a defendant offered testimony that he was committing a bank robbery in another state on the day the victim was killed. The testimony is: (A) admissible as not hearsay. (B) admissible as an admission. $\ldots$ \\
    \hdashline
    Medical & \multirow{2}{=}{Mixed task types: knowledge explanation vs.\ symptom-based consultation} &
    \textit{[Knowledge]} What does ``androgenic'' mean in the context of anabolic-androgenic steroids? \\
    \cdashline{3-3}
    & &
    \textit{[Consultation]} [Clinical case description: patient history, symptoms, treatment course$\ldots$] What clinical management approach is most appropriate? \\
    \hline
  \end{tabular}
  \caption{Representative instructions from each domain in the preliminary experiment.
  Code and history domains exhibit highly uniform task structures (programming syntax and passage-based MCQ, respectively), driving their pronounced separation in activation space.
  Mathematics, law, and medical domains exhibit greater task diversity and share a reliance on natural language analytical reasoning, leading to overlapping activation patterns and smaller inter-domain separation.}
  \label{tab:domain_examples}
\end{table*}

\end{document}